\documentclass{article} 
\usepackage{arxiv,times}

\usepackage{amsmath,amsfonts,bm}

\def\eqref#1{equation~\ref{#1}}

\def\1{\bm{1}}

\DeclareMathAlphabet{\mathsfit}{\encodingdefault}{\sfdefault}{m}{sl}
\SetMathAlphabet{\mathsfit}{bold}{\encodingdefault}{\sfdefault}{bx}{n}

\usepackage{hyperref}
\usepackage{url}
\usepackage{graphicx}
\usepackage{booktabs}
 \usepackage{multirow}

\title{CTIA-ODE: Continuous-Time Inter-Variable Attention for Long-Term Time Series Forecasting}

\iclrfinalcopy

\author{%
  Chengyi Pan \\
   HKUST-GZ \\
  \texttt{pancycypan@hkust-gz.edu.cn} \\
  \And
  Huangqimei Zheng \\
  School of Software, Yunnan University \\
  \texttt{zhenghuangqimei@stu.ynu.edu.cn} \\
  \And
  Hao Xue\thanks{Corresponding author.} \\
  The Hong Kong University of Science and Technology (Guangzhou) \\
  \texttt{haoxue@hkust-gz.edu.cn} \\
}

\begin{document}

\maketitle
\begin{abstract}
Although multivariate time series forecasting has broad application prospects, effectively incorporating exogenous variables remains challenging. Prevailing discrete-time architectures often impose rigid temporal alignment across heterogeneous variables, which may limit their ability to capture continuously evolving external effects. To address this limitation, we introduce CTIA-ODE, a continuous-time forecasting framework centered on Continuous-Time Inter-Variable Attention (CTIA). Specifically, CTIA is embedded within an ordinary differential equation (ODE) solver, enabling the continuously evolving endogenous trajectory to dynamically query a static exogenous context along the integration path without requiring explicit temporal alignment. To improve the stability of the continuous latent dynamics, we further introduce a dual-stabilization strategy: architecturally, a Linear Residual Augmentation (LRA) module provides a macroscopic trend anchor, while an Outlier-Robust Hybrid Loss (ORHL) mitigates the influence of heavy-tailed anomalies during optimization. Extensive experiments across eight real-world datasets demonstrate that CTIA-ODE achieves state-of-the-art forecasting performance across diverse datasets and prediction horizons, while maintaining a favorable accuracy-efficiency trade-off with a compact memory footprint and low inference latency, particularly under long-horizon forecasting settings. Anonymous code is available at \url{https://anonymous.4open.science/r/CTIA-ODE-15A6/}.
\end{abstract}

\section{Introduction}

Time series forecasting plays a crucial role in numerous key domains, including economics \citep{zhao2026survey}, weather evolution \citep{pan2026cityscalecast} and healthcare management \citep{che2025fusion}. In real-world complex systems, such as power grids or financial markets, the severe fluctuations of target endogenous variables are fundamentally driven by external factors like holidays, macroeconomic indicators \citep{carriero2024macroeconomic}, or extreme weather \citep{zhao2026survey,zhou2025crosslinear,dong2026ftsformer}. Ignoring these exogenous variables in forecasting models directly caps the predictive upper bound, as they cannot anticipate systemic state shifts brought by external shocks \citep{wang2024timexer,alaali2026beyond}.

\begin{figure*}[t]
\begin{center}
\includegraphics[width=0.98\linewidth]{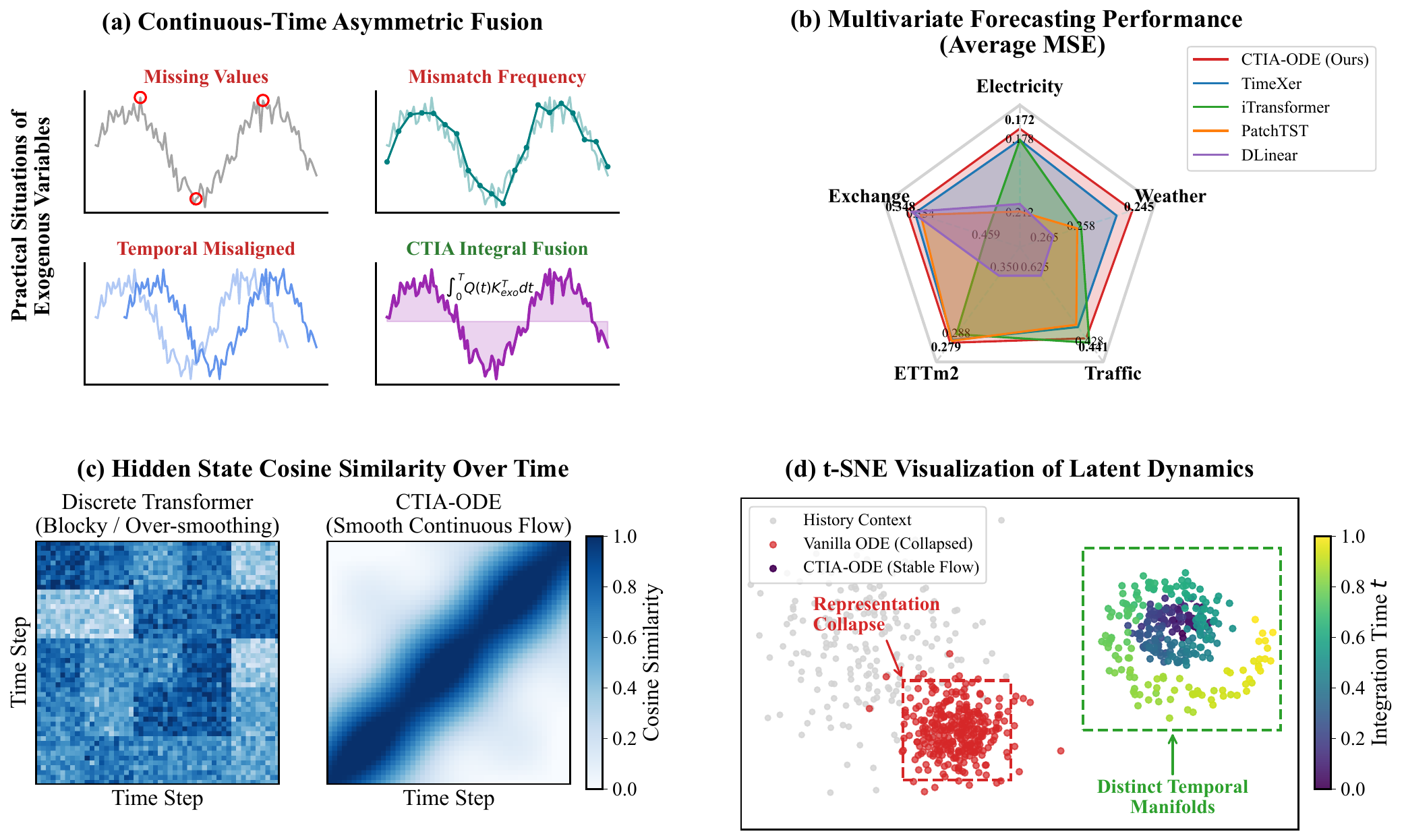}
\end{center}
\vspace{-1ex}\vspace{-3mm}
\caption{Motivation and empirical characteristics of CTIA-ODE. (a) Real-world exogenous variables may exhibit missing values, temporal misalignment, and frequency mismatch, motivating an asymmetric continuous-time fusion paradigm. (b) Forecasting performance on representative benchmarks. (c) Evolution of hidden-state similarity over the integration process, illustrating the difference between discrete representations and the continuous latent dynamics of CTIA-ODE. (d) t-SNE visualization of the learned latent states, providing a qualitative view of their temporal organization and representation stability.}
\label{fig:teaser}
\vspace{-3ex}
\end{figure*}

The mainstream approaches in recent years have mainly been based on Recurrent Neural Networks (RNNs) \citep{lin2025segrnn} and Transformers \citep{zhao2026survey,liu2025timer,chen2025closer}. RNNs rely on the step-by-step update of hidden states to process sequences, smoothly capturing dynamic temporal dependencies \citep{alharthi2024enhanced,sun2026review}. Transformer architectures utilize global self-attention mechanisms to achieve parallel computation across time steps, demonstrating exceptional expressive power in capturing long-range temporal correlations \citep{zhao2026survey,soe2026adaptive}. However, when attempting to simultaneously capture cross-correlations among multiple variables, both paradigms frequently fall into over-parameterization \citep{li2024improving}. Real-world data is often limited in samples and accompanied by unpredictable noise\citep{godahewa2021monash}; overly complex dependency modeling mechanisms and excessively deep network architectures easily lead to severe overfitting \citep{zhao2026survey,zhou2025crosslinear,huang2023crossgnn}. This not only degrades the model's generalization ability but also causes precious high-frequency dynamic information to be over-smoothed \citep{hu2026bridging}. This issue is not limited to prediction accuracy, but also concerns how latent representations evolve during forecasting. As illustrated in Figure \ref{fig:teaser}(c–d), we therefore examine both the temporal evolution and latent-space organization of hidden states, providing complementary evidence for the continuous dynamics modeled by CTIA-ODE.

Beyond latent representation dynamics, incorporating exogenous variables introduces another layer of difficulty. Exogenous and endogenous variables possess distinctly different physical meanings \citep{zhou2025crosslinear,bi2024spatiotemporal}. Conflating them indiscriminately not only creates massive computational redundancy but also induces unnecessary reverse causal interference \citep{cheong2026assessing,zhou2025crosslinear,wang2024timexer}. Therefore, striking a balance between endogenous and exogenous features is particularly important. In past deep learning research, the many-to-one forecasting paradigm with exogenous variables was insufficiently explored\citep{han2026causal,zhou2025crosslinear}, but recently this core pain point has rapidly become a major focus in the time series community \citep{zhou2025crosslinear}. Cutting-edge methods have proposed various fusion strategies. TimeXer employs an asymmetric representation granularity, utilizing global endogenous tokens to interact with exogenous variable tokens \citep{wang2024timexer}. CrossLinear introduces a cross-correlation embedding module to integrate external information \citep{zhou2025crosslinear}. ChronosX incorporates covariates into pretrained time series models \citep{cheong2026assessing}.


Despite significant progress, existing methods still face important limitations in incorporating exogenous information. To circumvent variable frequency mismatches, many approaches reduce the role of exogenous variables to simple auxiliary contextual representations \citep{liu2026global,wu2025millgnn,zhou2025crosslinear}. However, a more fundamental challenge arises from the discrete architectures commonly used for multivariate forecasting \citep{bartolucci2019hybrid,zhang2024asymmetric}. When heterogeneous endogenous and exogenous sequences are forced into strict step-by-step temporal alignment, substantial computational redundancy arises, while the resulting symmetric interactions frequently introduce unnecessary interference between primary and auxiliary variables \citep{yuan2026tackling,habiba2026emerging}. Consequently, conventional discrete attention mechanisms have difficulty modeling continuous interactions between an evolving endogenous trajectory and heterogeneous exogenous information without explicit temporal alignment \citep{wang2024timexer}.

Based on the above analysis, this study introduces a continuous-time perspective based on ordinary differential equations (ODEs) \citep{rubanova2019latent,kidger2020neural} and proposes CTIA-ODE, a continuous-time forecasting architecture centered on Continuous-Time Inter-Variable Attention (CTIA). The model evolves the latent endogenous representation through an ODE solver while allowing the continuously evolving trajectory to dynamically generate query representations along the integration path. These queries interact with a static exogenous context constructed from external variables, enabling continuous-time cross-variable interaction without explicit step-by-step temporal alignment. To enhance the stability of the resulting latent dynamics, CTIA-ODE further incorporates a Linear Residual Augmentation (LRA) module and an Outlier-Robust Hybrid Loss (ORHL). As illustrated in Figure \ref{fig:teaser}, this asymmetric design separates the evolving endogenous trajectory from the exogenous context while allowing information to flow continuously from the latter to the former.


Our primary contributions are summarized as follows:
\begin{itemize}
\item We propose CTIA-ODE, a continuous-time forecasting architecture centered on Continuous-Time Inter-Variable Attention (CTIA). This asymmetric mechanism enables the continuously evolving endogenous trajectory to dynamically query a static exogenous context, achieving cross-variable interaction without explicit temporal alignment or quadratic temporal complexity.
\item We introduce a dual-stabilization strategy to safeguard the continuous latent dynamics. Architecturally, a Linear Residual Augmentation (LRA) module provides a macroscopic trend anchor; during optimization, an Outlier-Robust Hybrid Loss (ORHL) suppresses gradient spikes and mitigates the influence of heavy-tailed anomalies, significantly improving the robustness of the learned vector field.
\item Comprehensive evaluations across eight real-world benchmarks demonstrate that CTIA-ODE achieves state-of-the-art performance. Beyond superior predictive accuracy, our framework exhibits a favorable accuracy-efficiency trade-off, operating with a compact memory footprint and low inference latency, particularly under extreme long-horizon settings.
\end{itemize}

\section{Related Work}
\label{sec:related_work}

\subsection{Continuous-Time Forecasting Models}
Recent advancements in multivariate time series forecasting have been largely driven by discrete-time deep architectures, including efficient Transformers \citep{zhou2021informer, wu2021autoformer, nie2023time, zhang2026autohformer} and linear decomposition frameworks \citep{zeng2023transformers, li2023revisiting}. While linear models efficiently extract macroscopic trends, these discrete paradigms lack continuous physical priors, making them susceptible to error compounding over extended horizons. Neural ODEs \citep{chen2018neural} address this by formulating hidden state transitions as continuous vector fields\citep{hasani2021liquid,gu2021efficiently}. However, applying classical Neural ODEs or recent state-space variants \citep{chen2026mode,dupont2019augmented} directly to multivariate forecasting often leads to representation collapse, as a single solver struggles to handle both global trends and high-frequency spatial dynamics. Unlike these approaches\citep{massaroli2020dissecting,dupont2019augmented}, CTIA-ODE explicitly decouples the learning process. By employing the LRA module to anchor macroscopic trends and utilizing zero-initialization for the continuous branch, our architecture maintains stable vector fields and prevents error amplification over long trajectories.

\subsection{Robust Multivariate Dependencies}
Balancing inter-series correlations with distributional robustness remains a primary challenge in multivariate forecasting\citep{chen2025multivariate,kim2025comprehensive}. Channel-dependent architectures \citep{zhang2023crossformer, liu2024itransformer} capture cross-variate interactions but are prone to negative transfer when integrating heterogeneous variables, a vulnerability that asymmetric paradigms \citep{wang2024timexer} aim to mitigate. Additionally, real-world time series frequently exhibit heavy-tailed distribution shifts and non-stationarity \citep{hu2026bridging,fan2023dish,liu2022attribution}. While instance normalization \citep{kim2021reversible} alleviates basic mean-variance drifts, standard quadratic loss functions over-penalize outliers\citep{shang2025cognos,gao2023normalized}, severely distorting the continuous vector fields in dynamical models\citep{oreshkin2019n,le2019shape}. To address these interconnected challenges, CTIA-ODE introduces an asymmetric CTIA mechanism that selectively absorbs exogenous contexts without inducing symmetric cross-talk. Furthermore, we employ the Outlier-Robust Hybrid Loss, which combines localized quadratic convergence with a bounded linear penalty for large errors. This formulation provides strictly bounded gradients, enhancing optimization robustness against heavy-tailed sensor anomalies and helping to mitigate solver instability.

\section{Methodology}
\subsection{Problem Formulation}

Given a historical observation sequence $\mathcal{X} \in \mathbb{R}^{M \times L}$, the multivariate forecasting objective is to predict future values for all variables $\mathcal{Y} \in \mathbb{R}^{M \times T}$ over a horizon $T$. To model asymmetric cross-variable interactions, we designate one feature as the primary variable and use the remaining variables as auxiliary contextual variables. The remaining \(M-1\) variables are encoded as a shared auxiliary context \(X_{ex}\in\mathbb{R}^{K_{ex}\times L}\), where \(K_{ex}=M-1\). The model learns a mapping function $\mathcal{F}_\theta$:
\begin{equation}
\hat{\mathcal{Y}} = \mathcal{F}_\theta(\mathcal{X}, X_{ex}, T_f),
\end{equation}
where $T_f$ represents the known future time features and $\theta$ denotes learnable parameters. During integration, all $M$ endogenous trajectories dynamically query this shared context, yielding the final multivariate prediction.

\subsection{Structure Overview}

\begin{figure}[t]
    \centering
    \includegraphics[width=\textwidth]{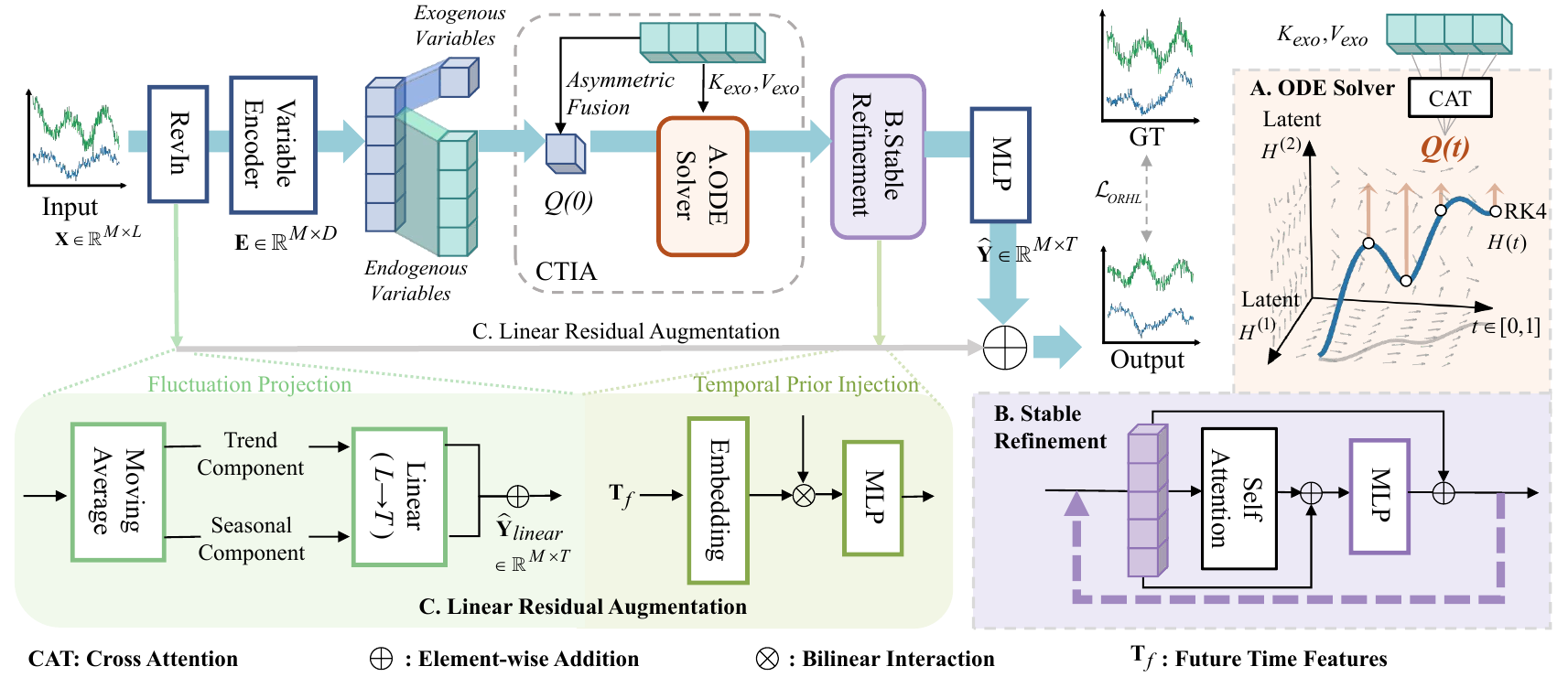}
    \caption{The overall framework of the proposed CTIA-ODE. (A) ODE Solver drives the continuous-time evolution of hidden states via RK4 integration, dynamically querying the static exogenous field through CTIA. (B) Stable Refinement performs discrete filtering on terminal states to maintain manifold stability. (C) Linear Residual Augmentation provides a robust prediction baseline by decoupling historical fluctuations and injecting future time priors.}
    \label{fig:overall_architecture}
    \vspace{-1ex}
\end{figure}

As illustrated in Figure \ref{fig:overall_architecture}, the main path of the proposed CTIA-ODE architecture is based on the continuous-time integration method. The raw input sequence $X \in \mathbb{R}^{M \times L}$ is first normalized via the RevIN module \citep{kim2021reversible}. Subsequently, a simple variable encoder maps the sequence into a high-dimensional latent space, yielding representations $E \in \mathbb{R}^{M \times D}$. These dense representations are then decoupled into endogenous variables and exogenous variables. 

In the core prediction workflow, the decoupled representations enter the CTIA module. An initial Asymmetric Fusion operates on these variables. Subsequently, the ODE Solver models the continuous-time evolution of the hidden state $H(t)$ over the integration interval $t \in [0,1]$. The terminal state derived from the ODE solver is then fed into a Stable Refinement module. This block stabilizes the latent manifold and filters integration noise before passing the representations through the final Multi-Layer Perceptron (MLP). Operating in parallel, the LRA module provides prediction enhancement by performing two different linear interaction mechanisms on distinct inputs. Finally, the outputs from both pathways are aggregated via addition and subsequently denormalized to produce the ultimate prediction $\hat{Y}$. The entire framework is optimized end-to-end guided by the Outlier-Robust Hybrid Loss ($\mathcal{L}_{ORHL}$).

\subsection{Continuous-Time Integral Attention}

We formulate the temporal evolution of the endogenous variables as a continuous-time dynamical system. To seamlessly incorporate the forcing exerted by exogenous variables, we propose CTIA. Unlike standard discrete attention applied at isolated network layers, CTIA evaluates cross-variable dependencies continuously along the integration trajectory.

Let $E_{en}$ and $E_{ex}$ denote the encoded latent representations of the endogenous and exogenous variables, respectively. Prior to the ODE integration, we project $E_{ex}$ into a time-invariant exogenous field, obtaining the static key and value representations:
\begin{equation}
    K_{exo} = E_{ex}W_K, \quad V_{exo} = E_{ex}W_V,
\end{equation}
where $W_K, W_V \in \mathbb{R}^{D \times d_k}$ are learnable projection weight matrices. Before entering the continuous solver, an initial Asymmetric Fusion is applied to construct the starting state of the endogenous system $H(0)$ by fusing $E_{en}$ with the static exogenous field.

During the continuous ODE integration spanning $t \in [0,1]$, the endogenous hidden state $H(t)$ acts as a moving agent. At any instantaneous time step $t$, it generates a dynamic query $Q(t) = H(t)W_Q$. The instantaneous exogenous forcing $\mathcal{C}(t)$ is thus computed via a continuous cross-attention mechanism:
\begin{equation}
    \mathcal{C}(t) = \text{Softmax}\left(\frac{Q(t)K_{exo}^\top}{\sqrt{d_k}}\right)V_{exo}.
\end{equation}

Ultimately, the derivative governing the evolution of the endogenous state is defined by superimposing the base dynamics with the dynamic exogenous forcing:
\begin{equation}
    \frac{dH(t)}{dt} = f_{\phi}(H(t)) + \alpha \mathcal{C}(t),
\end{equation}
where $f_{\phi}(\cdot)$ is a MLP parameterizing the base vector field, and $\alpha$ is a learnable scaling parameter. In this formulation, $K_{exo}$ and $V_{exo}$ provide a stable global context, while $Q(t)$ continuously queries this field to adapt the evolutionary trajectory of $H(t)$.

\subsection{Stable Refinement}

Following the ODE integration, the terminal latent state $H(1)$ inevitably accumulates numerical noise. To filter this noise while preserving critical temporal details, we introduce the weight-sharing Dual-Loop Stable Refinement module.

As illustrated in Figure \ref{fig:overall_architecture}(B), taking $H(1)$ as the initial input, the refinement process iteratively applies a Self-Attention layer and a MLP, both equipped with residual connections and layer normalization. We utilize a weight-sharing recurrent design constrained to exactly two iterations. This dual-loop mechanism acts as a latent double-check while keeping the parameter footprint to a minimum.

\subsection{Linear Residual Augmentation}

This auxiliary module consists of two specific mechanisms. First, the Fluctuation Projection takes the normalized historical sequence $\mathbf{X}_{norm}$ as input. It decouples the raw sequence into trend and seasonal components via a moving average operation. These two components are then independently mapped from the look-back window $L$ to the forecasting horizon $T$ through linear layers and aggregated. 

Second, the Temporal Prior Injection takes both the known future time features $\mathbf{T}_f$ and the refined terminal state $\tilde{H}_{sr}^{(2)}$ (from the Stable Refinement block) as inputs. It encodes $\mathbf{T}_f$ and fuses it with the refined latent manifold via a bilinear interaction mechanism to yield auxiliary temporal effects. Finally, the outputs from these two parallel mechanisms are superimposed onto the primary non-linear prediction via learnable scalar weights to form the ultimate forecast.

\subsection{Outlier-Robust Hybrid Loss}
Real-world physical sensor data often contains non-stationary distributions and outliers. Training continuous dynamical systems under standard quadratic objectives can heavily penalize these anomalies, inducing large gradients that distort the learned vector field. To improve robustness, we optimize the aggregated final prediction $\hat{Y}$ using the Outlier-Robust Hybrid Loss ($\mathcal{L}_{\text{ORHL}}$). Given the ground truth $Y$, let $e = \hat{Y} - Y$ denote the forecasting residual. The ORHL applies a piecewise penalty function element-wise:
\begin{equation}
    \mathcal{L}_{\text{ORHL}}(e) = \frac{1}{N} \sum_{i=1}^{N} 
    \begin{cases} 
        \frac{1}{2} e_i^2 + \lambda |e_i|, & \text{if } |e_i| \le \delta, \\
        (\delta + \lambda) |e_i| - \frac{1}{2} \delta^2, & \text{otherwise},
    \end{cases}
\end{equation}
where $\delta$ dictates the transition threshold and $\lambda$ acts as a linear scaling factor. 

This formulation provides two critical properties for continuous-time modeling. In the localized convergence zone ($|e_i| \le \delta$), the objective operates as an Elastic-Net-style composite: the quadratic curvature provides stable optimization, while the scaled $\ell_1$ penalty encourages sparsity in the residual distribution. For large anomalous deviations ($|e_i| > \delta$), the gradient strictly saturates at a bounded constant $\pm(\delta + \lambda)$. This bounded prediction gradient limits the magnitude of loss signals entering the continuous branch and improves optimization robustness under large residuals.

\section{Experiments}
\label{sec:experiments}

We evaluate CTIA-ODE on multiple real-world multivariate time series forecasting benchmarks. The experiments are designed to answer the following research questions: \textbf{RQ1:} How does CTIA-ODE compare against state-of-the-art (SOTA) forecasters across varying prediction horizons? \textbf{RQ2:} What are the individual contributions of the Outlier-Robust Hybrid Loss (ORHL), the LRA module, and the continuous-time attention mechanism? \textbf{RQ3:} Does the model achieve a favorable trade-off between computational efficiency and forecasting robustness?

\subsection{Experimental Settings}

\textbf{Datasets.} We conduct experiments on eight widely used real-world multivariate benchmarks: Electricity, Exchange, Traffic, Weather, and the ETT datasets (ETTh1, ETTh2, ETTm1, ETTm2). These datasets encompass diverse domains, ranging from energy management to meteorological forecasting, with varying periodicities and dimensionalities.

\textbf{Baselines and Evaluation Metrics.} We comprehensively evaluate CTIA-ODE against nine state-of-the-art multivariate forecasting models, which can be categorized into four distinct architectural paradigms: (1) Transformer-based models: TimeXer \citep{wang2024timexer}, iTransformer \citep{liu2024itransformer}, PatchTST \citep{nie2023time}, Crossformer \citep{zhang2023crossformer}, and Autoformer \citep{wu2021autoformer}; (2) State-Space Models (SSMs): S-Mamba\citep{wang2025mamba}; (3) Linear models: DLinear \citep{zeng2023transformers}; (4) CNN and Time-Frequency architectures: TimesNet \citep{wu2023timesnet} and WaveTS-B\citep{zhou2026wavelet}.
To quantitatively evaluate the forecasting performance, we use the standard Mean Squared Error (MSE) and Mean Absolute Error (MAE). 


\textbf{Implementation Details.} Following the standard protocol in Multivariate Time Series Forecasting, the look-back window length $L$ is fixed to 96, and the prediction horizon $T$ is evaluated at $\{96, 192, 336, 720\}$. We utilize the AdamW optimizer with an initial learning rate of $5\times 10^{-4}$ and a weight decay of $1\times 10^{-4}$. The learning rate is dynamically adjusted using a ReduceLROnPlateau scheduler with a decay factor of 0.5 and a patience of 3 epochs, bounded by a minimum learning rate of $1\times 10^{-6}$. To strike an optimal balance between representational capacity and parameter efficiency, the hidden dimension d\_{model} is set to 128, completely avoiding the heavy memory overhead of typical 512-dimensional Transformers. The network is trained using our proposed ORHL with an early stopping patience of 12 epochs.

\begin{table}[htb]
\vspace{-2ex}
\caption{Multivariate forecasting results. The look-back length $L$ is set to 96. The best results are highlighted in \textbf{bold} and the second best are \underline{underlined}.}
\label{tab:main_results}
\begin{center}
\resizebox{\textwidth}{!}{
\begin{tabular}{cc|cc|cc|cc|cc|cc|cc|cc|cc|cc|cc}
\toprule
\multicolumn{2}{c|}{\multirow{2}{*}{Models}} & \multicolumn{2}{c|}{\textbf{CTIA-ODE (Ours)}} & \multicolumn{2}{c|}{WaveTS-B} & \multicolumn{2}{c|}{S-Mamba} & \multicolumn{2}{c|}{TimeXer} & \multicolumn{2}{c|}{iTransformer} & \multicolumn{2}{c|}{PatchTST} & \multicolumn{2}{c|}{DLinear} & \multicolumn{2}{c|}{Crossformer} & \multicolumn{2}{c|}{TimesNet} & \multicolumn{2}{c}{Autoformer} \\
\cmidrule(lr){3-4} \cmidrule(lr){5-6} \cmidrule(lr){7-8} \cmidrule(lr){9-10} \cmidrule(lr){11-12} \cmidrule(lr){13-14} \cmidrule(lr){15-16} \cmidrule(lr){17-18} \cmidrule(lr){19-20} \cmidrule(lr){21-22}
\multicolumn{2}{c|}{Metric} & MSE & MAE & MSE & MAE & MSE & MAE & MSE & MAE & MSE & MAE & MSE & MAE & MSE & MAE & MSE & MAE & MSE & MAE & MSE & MAE \\
\midrule

\multirow{5}{*}{\rotatebox{90}{\textbf{Electricity}}} 
& 96  & \textbf{0.140} & \textbf{0.233} & \underline{0.143} & \underline{0.235} & 0.147 & 0.249 & 0.147 & 0.251 & 0.148 & 0.240 & 0.195 & 0.285 & 0.197 & 0.282 & 0.219 & 0.314 & 0.168 & 0.272 & 0.201 & 0.317 \\ 
& 192 & \textbf{0.160} & \textbf{0.252} & 0.164 & 0.255 & 0.167 & 0.270 & 0.163 & 0.266 & \underline{0.162} & \underline{0.253} & 0.199 & 0.289 & 0.196 & 0.285 & 0.231 & 0.322 & 0.184 & 0.289 & 0.222 & 0.334 \\ 
& 336 & \underline{0.181} & \underline{0.274} & 0.185 & 0.278 & 0.189 & 0.291 & 0.184 & 0.282 & \textbf{0.178} & \textbf{0.269} & 0.215 & 0.305 & 0.209 & 0.301 & 0.246 & 0.337 & 0.198 & 0.300 & 0.231 & 0.338 \\ 
& 720 & \textbf{0.205} & \textbf{0.296} & \underline{0.211} & \underline{0.302} & 0.232 & 0.327 & 0.220 & 0.312 & 0.225 & 0.317 & 0.256 & 0.337 & 0.245 & 0.333 & 0.280 & 0.363 & 0.220 & 0.320 & 0.254 & 0.361 \\ 
\cmidrule(lr){2-22}
& Avg & \textbf{0.172} & \textbf{0.264} & \underline{0.176} & \underline{0.268} & 0.183 & 0.284 & 0.178 & 0.278 & 0.178 & 0.270 & 0.216 & 0.304 & 0.212 & 0.300 & 0.244 & 0.334 & 0.192 & 0.295 & 0.227 & 0.338 \\ 
\midrule

\multirow{5}{*}{\rotatebox{90}{\textbf{Exchange}}} 
& 96  & 0.093 & 0.215 & 0.098 & 0.219 & \underline{0.089} & \underline{0.211} & 0.101 & 0.226 & 0.118 & 0.253 & \textbf{0.088} & \textbf{0.205} & \textbf{0.088} & 0.218 & 0.256 & 0.367 & 0.107 & 0.234 & 0.197 & 0.323 \\ 
& 192 & 0.188 & \underline{0.308} & 0.192 & 0.311 & \underline{0.187} & 0.309 & 0.199 & 0.317 & 0.232 & 0.355 & \textbf{0.176} & \textbf{0.299} & \textbf{0.176} & 0.315 & 0.470 & 0.509 & 0.226 & 0.344 & 0.300 & 0.369 \\ 
& 336 & 0.325 & \underline{0.415} & 0.342 & 0.428 & 0.338 & 0.423 & 0.334 & 0.424 & 0.434 & 0.489 & \textbf{0.301} & \textbf{0.397} & \underline{0.313} & 0.427 & 1.268 & 0.883 & 0.367 & 0.448 & 0.509 & 0.524 \\ 
& 720 & \textbf{0.788} & \textbf{0.668} & 1.031 & 0.768 & 0.875 & 0.710 & \underline{0.803} & \underline{0.680} & 1.050 & 0.750 & 0.901 & 0.714 & 0.839 & 0.695 & 1.767 & 1.068 & 0.964 & 0.746 & 1.447 & 0.941 \\ 
\cmidrule(lr){2-22}
& Avg & \textbf{0.348} & \textbf{0.402} & 0.415 & 0.431 & 0.372 & 0.413 & 0.359 & 0.411 & 0.459 & 0.463 & 0.367 & \underline{0.404} & \underline{0.354} & 0.414 & 0.940 & 0.707 & 0.416 & 0.443 & 0.613 & 0.539 \\ 
\midrule

\multirow{5}{*}{\rotatebox{90}{\textbf{Traffic}}} 
& 96  & \underline{0.411} & \textbf{0.253} & 0.416 & 0.271 & 0.428 & 0.271 & 0.435 & 0.279 & \textbf{0.395} & \underline{0.268} & 0.462 & 0.295 & 0.650 & 0.396 & 0.522 & 0.290 & 0.593 & 0.321 & 0.613 & 0.388 \\ 
& 192 & \underline{0.425} & \textbf{0.262} & 0.431 & 0.278 & 0.442 & 0.280 & 0.457 & 0.288 & \textbf{0.417} & \underline{0.276} & 0.466 & 0.296 & 0.598 & 0.370 & 0.530 & 0.293 & 0.617 & 0.336 & 0.616 & 0.382 \\ 
& 336 & \underline{0.456} & \textbf{0.270} & 0.462 & \underline{0.281} & 0.471 & 0.292 & 0.483 & 0.296 & \textbf{0.433} & 0.283 & 0.482 & 0.304 & 0.605 & 0.373 & 0.558 & 0.305 & 0.629 & 0.336 & 0.622 & 0.337 \\ 
& 720 & \underline{0.474} & \textbf{0.285} & 0.481 & \underline{0.302} & 0.495 & 0.308 & 0.522 & 0.316 & \textbf{0.467} & \underline{0.302} & 0.514 & 0.322 & 0.645 & 0.394 & 0.589 & 0.328 & 0.640 & 0.350 & 0.660 & 0.408 \\ 
\cmidrule(lr){2-22}
& Avg & \underline{0.441} & \textbf{0.268} & 0.448 & 0.283 & 0.459 & 0.288 & 0.474 & 0.293 & \textbf{0.428} & \underline{0.282} & 0.481 & 0.304 & 0.625 & 0.383 & 0.550 & 0.304 & 0.620 & 0.336 & 0.628 & 0.379 \\ 
\midrule

\multirow{5}{*}{\rotatebox{90}{\textbf{Weather}}} 
& 96  & \textbf{0.153} & \textbf{0.191} & 0.172 & 0.225 & 0.169 & \underline{0.210} & 0.165 & 0.214 & 0.174 & 0.214 & 0.177 & 0.218 & 0.196 & 0.255 & \underline{0.158} & 0.230 & 0.172 & 0.220 & 0.266 & 0.336 \\ 
& 192 & \underline{0.208} & \textbf{0.247} & 0.215 & 0.261 & 0.214 & \underline{0.253} & 0.211 & \underline{0.253} & 0.221 & 0.254 & 0.225 & 0.259 & 0.237 & 0.296 & \textbf{0.206} & 0.277 & 0.219 & 0.261 & 0.307 & 0.367 \\ 
& 336 & \textbf{0.267} & \textbf{0.288} & 0.273 & 0.298 & 0.274 & \underline{0.296} & \underline{0.271} & 0.298 & 0.278 & \underline{0.296} & 0.278 & 0.297 & 0.283 & 0.335 & 0.272 & 0.335 & 0.280 & 0.306 & 0.359 & 0.395 \\ 
& 720 & 0.351 & \textbf{0.342} & 0.358 & 0.352 & 0.353 & \underline{0.348} & \underline{0.346} & \underline{0.348} & 0.358 & 0.349 & 0.354 & \underline{0.348} & \textbf{0.345} & 0.381 & 0.398 & 0.418 & 0.365 & 0.359 & 0.419 & 0.428 \\ 
\cmidrule(lr){2-22}
& Avg & \textbf{0.245} & \textbf{0.267} & 0.254 & 0.284 & 0.253 & \underline{0.277} & \underline{0.249} & 0.280 & 0.258 & 0.279 & 0.259 & 0.281 & 0.265 & 0.317 & 0.259 & 0.315 & 0.259 & 0.287 & 0.338 & 0.382 \\ 
\midrule

\multirow{5}{*}{\rotatebox{90}{\textbf{ETTh1}}} 
& 96  & 0.390 & \textbf{0.397} & 0.396 & 0.406 & \underline{0.386} & 0.405 & 0.391 & 0.409 & \underline{0.386} & 0.405 & 0.414 & 0.419 & \underline{0.386} & \underline{0.400} & 0.423 & 0.448 & \textbf{0.384} & 0.402 & 0.449 & 0.459 \\ 
& 192 & 0.455 & 0.437 & 0.462 & 0.443 & 0.443 & 0.437 & \textbf{0.436} & 0.445 & 0.441 & 0.436 & 0.460 & 0.445 & \underline{0.437} & \underline{0.432} & 0.471 & 0.474 & \textbf{0.436} & \textbf{0.429} & 0.500 & 0.482 \\ 
& 336 & 0.503 & 0.464 & 0.511 & 0.472 & 0.489 & 0.468 & \textbf{0.476} & \textbf{0.455} & 0.487 & \underline{0.458} & 0.501 & 0.466 & \underline{0.481} & 0.459 & 0.570 & 0.546 & 0.491 & 0.469 & 0.521 & 0.496 \\ 
& 720 & 0.526 & 0.504 & 0.534 & 0.512 & 0.502 & 0.489 & \textbf{0.475} & \textbf{0.470} & 0.503 & 0.491 & \underline{0.500} & \underline{0.488} & 0.519 & 0.516 & 0.653 & 0.621 & 0.521 & 0.500 & 0.514 & 0.512 \\ 
\cmidrule(lr){2-22}
& Avg & 0.469 & 0.450 & 0.476 & 0.458 & 0.455 & 0.450 & \textbf{0.445} & \textbf{0.444} & \underline{0.454} & \underline{0.447} & 0.469 & 0.454 & 0.456 & 0.452 & 0.529 & 0.522 & 0.458 & 0.450 & 0.496 & 0.487 \\ 
\midrule

\multirow{5}{*}{\rotatebox{90}{\textbf{ETTh2}}} 
& 96  & \textbf{0.290} & \textbf{0.338} & 0.295 & \underline{0.344} & 0.296 & 0.348 & \underline{0.293} & 0.347 & 0.297 & 0.349 & 0.302 & 0.348 & 0.333 & 0.387 & 0.745 & 0.584 & 0.340 & 0.374 & 0.346 & 0.388 \\ 
& 192 & 0.383 & \underline{0.396} & 0.389 & 0.403 & \underline{0.376} & \underline{0.396} & \textbf{0.373} & \textbf{0.395} & 0.380 & 0.400 & 0.388 & 0.400 & 0.477 & 0.476 & 0.877 & 0.656 & 0.402 & 0.414 & 0.456 & 0.452 \\ 
& 336 & 0.428 & 0.432 & 0.435 & 0.441 & \underline{0.424} & \underline{0.431} & \textbf{0.422} & \textbf{0.430} & 0.428 & 0.432 & 0.426 & 0.433 & 0.594 & 0.541 & 1.043 & 0.731 & 0.452 & 0.452 & 0.482 & 0.486 \\ 
& 720 & 0.433 & \underline{0.444} & 0.441 & 0.452 & \underline{0.426} & \underline{0.444} & \textbf{0.414} & \textbf{0.442} & 0.427 & 0.445 & 0.431 & 0.446 & 0.831 & 0.657 & 1.104 & 0.763 & 0.462 & 0.468 & 0.515 & 0.511 \\ 
\cmidrule(lr){2-22}
& Avg & 0.383 & \textbf{0.402} & 0.390 & 0.410 & \underline{0.381} & 0.405 & \textbf{0.375} & \underline{0.404} & 0.383 & 0.407 & 0.387 & 0.407 & 0.559 & 0.515 & 0.942 & 0.684 & 0.414 & 0.427 & 0.450 & 0.459 \\ 
\midrule

\multirow{5}{*}{\rotatebox{90}{\textbf{ETTm1}}} 
& 96  & \textbf{0.326} & \textbf{0.351} & 0.333 & \underline{0.358} & 0.333 & 0.368 & \underline{0.327} & 0.363 & 0.334 & 0.368 & 0.329 & 0.367 & 0.345 & 0.372 & 0.404 & 0.426 & 0.338 & 0.375 & 0.505 & 0.475 \\ 
& 192 & 0.378 & \textbf{0.377} & 0.385 & \underline{0.384} & 0.376 & 0.390 & \underline{0.370} & 0.389 & 0.377 & 0.391 & \textbf{0.367} & 0.385 & 0.380 & 0.389 & 0.450 & 0.451 & 0.374 & 0.387 & 0.553 & 0.496 \\ 
& 336 & 0.408 & \textbf{0.399} & 0.415 & \underline{0.406} & 0.408 & 0.413 & \underline{0.405} & 0.415 & 0.426 & 0.420 & \textbf{0.399} & 0.410 & 0.413 & 0.413 & 0.532 & 0.515 & 0.410 & 0.411 & 0.621 & 0.537 \\ 
& 720 & 0.472 & \textbf{0.434} & 0.481 & 0.442 & 0.475 & 0.448 & \underline{0.459} & 0.450 & 0.491 & 0.459 & \textbf{0.454} & \underline{0.439} & 0.474 & 0.453 & 0.666 & 0.589 & 0.478 & 0.450 & 0.671 & 0.561 \\ 
\cmidrule(lr){2-22}
& Avg & 0.396 & \textbf{0.390} & 0.403 & \underline{0.397} & 0.398 & 0.405 & \underline{0.390} & 0.405 & 0.407 & 0.410 & \textbf{0.387} & 0.400 & 0.403 & 0.407 & 0.513 & 0.496 & 0.400 & 0.406 & 0.588 & 0.517 \\ 
\midrule

\multirow{5}{*}{\rotatebox{90}{\textbf{ETTm2}}} 
& 96  & \textbf{0.175} & \textbf{0.253} & 0.181 & 0.260 & 0.179 & 0.263 & \underline{0.177} & 0.266 & 0.180 & 0.264 & \textbf{0.175} & \underline{0.259} & 0.193 & 0.292 & 0.287 & 0.366 & 0.187 & 0.267 & 0.255 & 0.339 \\ 
& 192 & \textbf{0.241} & \textbf{0.297} & 0.248 & 0.305 & 0.250 & 0.309 & \underline{0.245} & 0.306 & 0.250 & 0.309 & \textbf{0.241} & \underline{0.302} & 0.284 & 0.362 & 0.414 & 0.492 & 0.249 & 0.309 & 0.281 & 0.340 \\ 
& 336 & \textbf{0.300} & \textbf{0.334} & 0.308 & \underline{0.342} & 0.312 & 0.349 & \underline{0.305} & 0.344 & 0.311 & 0.348 & \underline{0.305} & 0.343 & 0.369 & 0.427 & 0.597 & 0.542 & 0.321 & 0.351 & 0.339 & 0.372 \\ 
& 720 & \underline{0.401} & \textbf{0.394} & 0.411 & 0.402 & 0.411 & 0.406 & \textbf{0.399} & 0.402 & 0.412 & 0.407 & 0.402 & \underline{0.400} & 0.554 & 0.522 & 1.730 & 1.042 & 0.408 & 0.403 & 0.433 & 0.432 \\ 
\cmidrule(lr){2-22}
& Avg & \textbf{0.279} & \textbf{0.319} & 0.287 & 0.327 & 0.288 & 0.332 & 0.282 & 0.331 & 0.288 & 0.332 & \underline{0.281} & \underline{0.326} & 0.350 & 0.401 & 0.757 & 0.610 & 0.291 & 0.333 & 0.327 & 0.371 \\ 
\midrule

\multicolumn{2}{c|}{\textbf{1$^{st}$ Count}} & \multicolumn{2}{c|}{\textbf{44}} & \multicolumn{2}{c|}{0} & \multicolumn{2}{c|}{0} & \multicolumn{2}{c|}{\underline{15}} & \multicolumn{2}{c|}{7} & \multicolumn{2}{c|}{12} & \multicolumn{2}{c|}{3} & \multicolumn{2}{c|}{1} & \multicolumn{2}{c|}{3} & \multicolumn{2}{c}{0} \\
\midrule
\multicolumn{2}{c|}{\textbf{Avg 1$^{st}$ Count}} & \multicolumn{2}{c|}{\textbf{11}} & \multicolumn{2}{c|}{0} & \multicolumn{2}{c|}{0} & \multicolumn{2}{c|}{\underline{3}} & \multicolumn{2}{c|}{1} & \multicolumn{2}{c|}{1} & \multicolumn{2}{c|}{0} & \multicolumn{2}{c|}{0} & \multicolumn{2}{c|}{0} & \multicolumn{2}{c}{0} \\
\bottomrule
\end{tabular}
}
\end{center}
\vspace{-3ex}
\end{table}

\subsection{Main Results (RQ1)}

Table \ref{tab:main_results} summarizes the multivariate forecasting performance across 8 representative benchmarks. A lower MSE or MAE indicates better predictive accuracy. 

As shown in the table, our CTIA-ODE achieves the best performance on 44 out of the evaluated metrics, consistently outperforming the leading baselines TimeXer (15 out of 80) and PatchTST (12 out of 80). A key advantage of CTIA-ODE is its robustness under extreme long-term horizons (e.g., $T=720$), where discrete-time models typically suffer from severe error accumulation. For example, on the ETTm1 dataset, CTIA-ODE maintains an MSE of 0.472 at $T=720$ compared to 0.326 at $T=96$, demonstrating a highly constrained error growth. This long-term stability across benchmarks like Electricity and ETTm2 validates our Augmented Continuous-Time Architecture: the LRA provides a macroscopic trend baseline to mitigate representation collapse, while the Neural ODE branch focuses on resolving local continuous dynamics.

Furthermore, CTIA-ODE effectively mitigates the negative transfer issue commonly observed in highly heterogeneous datasets (e.g., Exchange, Weather, and Traffic). While fully channel-dependent models such as Crossformer often suffer from non-causal cross-talk in these scenarios, CTIA-ODE addresses this through Continuous-Time Inter-Variable Attention. By processing exogenous variables asymmetrically as continuous contextual signals rather than computing symmetrical dense cross-attention, the model incorporates valid external constraints while filtering heterogeneous noise. Consequently, CTIA-ODE achieves the lowest MAE scores across all prediction horizons on the high-dimensional Traffic dataset, confirming its pointwise predictive reliability under complex spatial conditions.




\subsection{Ablation Study (RQ2)}

To evaluate the individual contributions of the proposed CTIA-ODE components, we conduct a detailed ablation study focusing on the Augmented Continuous-Time Architecture, the Continuous-Time Inter-Variable Attention, and the ORHL. We compare the full model against five ablated variants on the Electricity, ETTm1, Traffic, and Weather datasets. 

Table \ref{tab:ablation} presents the quantitative results across short ($T=96$) and extended ($T=720$) forecasting horizons. To validate the superiority of ORHL, we replace it with standard pure MSE and pure MAE losses. Predictive performance degrades under both single-metric configurations: pure MSE over-penalizes heavy-tailed anomalies, which distorts the continuous vector field, while pure MAE leads to slower convergence and suboptimal global optima. The proposed hybrid objective effectively mitigates both issues. 

Furthermore, to analyze the augmented architecture, we decouple the framework into a Pure ODE variant (removing the LRA) and a Pure Linear variant (removing the continuous dynamics). The Pure Linear variant struggles to capture high-frequency complex variations, yielding consistently higher errors. Conversely, while the Pure ODE variant performs adequately on short horizons, it suffers from severe representation collapse at $T=720$, evidenced by a nearly 16\% increase in MSE on the ETTm1 dataset. This demonstrates that the LRA module is crucial for maintaining macroscopic trend stability, while the ODE branch is necessary for resolving local microscopic dynamics. Finally, removing the exogenous variable interactions to enforce absolute channel independence results in noticeable performance drops, particularly on heterogeneous datasets like Traffic. This confirms that the asymmetric continuous cross-attention successfully captures multivariate correlations without inducing negative transfer.

\begin{table}
\vspace{-2ex}
\caption{Ablation study on Electricity, ETTm1, Traffic, and Weather datasets. The complete CTIA-ODE is compared against variants ablating the loss function, architecture, and attention mechanism. The best results are highlighted in \textbf{bold}.}
\label{tab:ablation}
\begin{center}
\resizebox{\textwidth}{!}{
\begin{tabular}{l|cc|cc|cc|cc|cc|cc|cc|cc}
\toprule
\multirow{3}{*}{Models} & \multicolumn{4}{c|}{Electricity} & \multicolumn{4}{c|}{ETTm1} & \multicolumn{4}{c|}{Traffic} & \multicolumn{4}{c}{Weather} \\
\cmidrule(lr){2-5} \cmidrule(lr){6-9} \cmidrule(lr){10-13} \cmidrule(lr){14-17}
& \multicolumn{2}{c|}{$T=96$} & \multicolumn{2}{c|}{$T=720$} & \multicolumn{2}{c|}{$T=96$} & \multicolumn{2}{c|}{$T=720$} & \multicolumn{2}{c|}{$T=96$} & \multicolumn{2}{c|}{$T=720$} & \multicolumn{2}{c|}{$T=96$} & \multicolumn{2}{c}{$T=720$} \\
\cmidrule(lr){2-3} \cmidrule(lr){4-5} \cmidrule(lr){6-7} \cmidrule(lr){8-9} \cmidrule(lr){10-11} \cmidrule(lr){12-13} \cmidrule(lr){14-15} \cmidrule(lr){16-17}
& MSE & MAE & MSE & MAE & MSE & MAE & MSE & MAE & MSE & MAE & MSE & MAE & MSE & MAE & MSE & MAE \\
\midrule
\textbf{CTIA-ODE (Full)} & \textbf{0.140} & \textbf{0.233} & \textbf{0.205} & \textbf{0.296} & \textbf{0.326} & \textbf{0.351} & \textbf{0.472} & \textbf{0.434} & \textbf{0.411} & \textbf{0.253} & \textbf{0.474} & \textbf{0.285} & \textbf{0.153} & \textbf{0.191} & \textbf{0.351} & \textbf{0.342} \\
\midrule
w/o Linear (Pure ODE) & 0.145 & 0.238 & 0.238 & 0.318 & 0.332 & 0.358 & 0.548 & 0.485 & 0.415 & 0.258 & 0.535 & 0.315 & 0.159 & 0.201 & 0.395 & 0.380 \\
w/o ODE (Pure Linear) & 0.149 & 0.245 & 0.215 & 0.302 & 0.341 & 0.365 & 0.488 & 0.448 & 0.430 & 0.270 & 0.492 & 0.300 & 0.165 & 0.208 & 0.368 & 0.355 \\
w/o CTIA (Pure CI) & 0.144 & 0.238 & 0.212 & 0.301 & 0.336 & 0.360 & 0.486 & 0.445 & 0.428 & 0.265 & 0.505 & 0.304 & 0.163 & 0.205 & 0.370 & 0.362 \\
w/o ORHL (Pure MSE) & 0.143 & 0.241 & 0.211 & 0.308 & 0.338 & 0.368 & 0.491 & 0.456 & 0.418 & 0.263 & 0.488 & 0.296 & 0.160 & 0.207 & 0.364 & 0.358 \\
w/o ORHL (Pure MAE) & 0.146 & 0.235 & 0.215 & 0.299 & 0.342 & 0.355 & 0.495 & 0.440 & 0.422 & 0.256 & 0.491 & 0.288 & 0.164 & 0.199 & 0.369 & 0.350 \\
\bottomrule
\end{tabular}
}
\end{center}
\vspace{-3ex}
\end{table}

\subsection{Model Efficiency and In-Depth Mechanistic Analysis (RQ3)}

\begin{figure*}[htb]
\begin{center}
\includegraphics[width=0.9\linewidth]{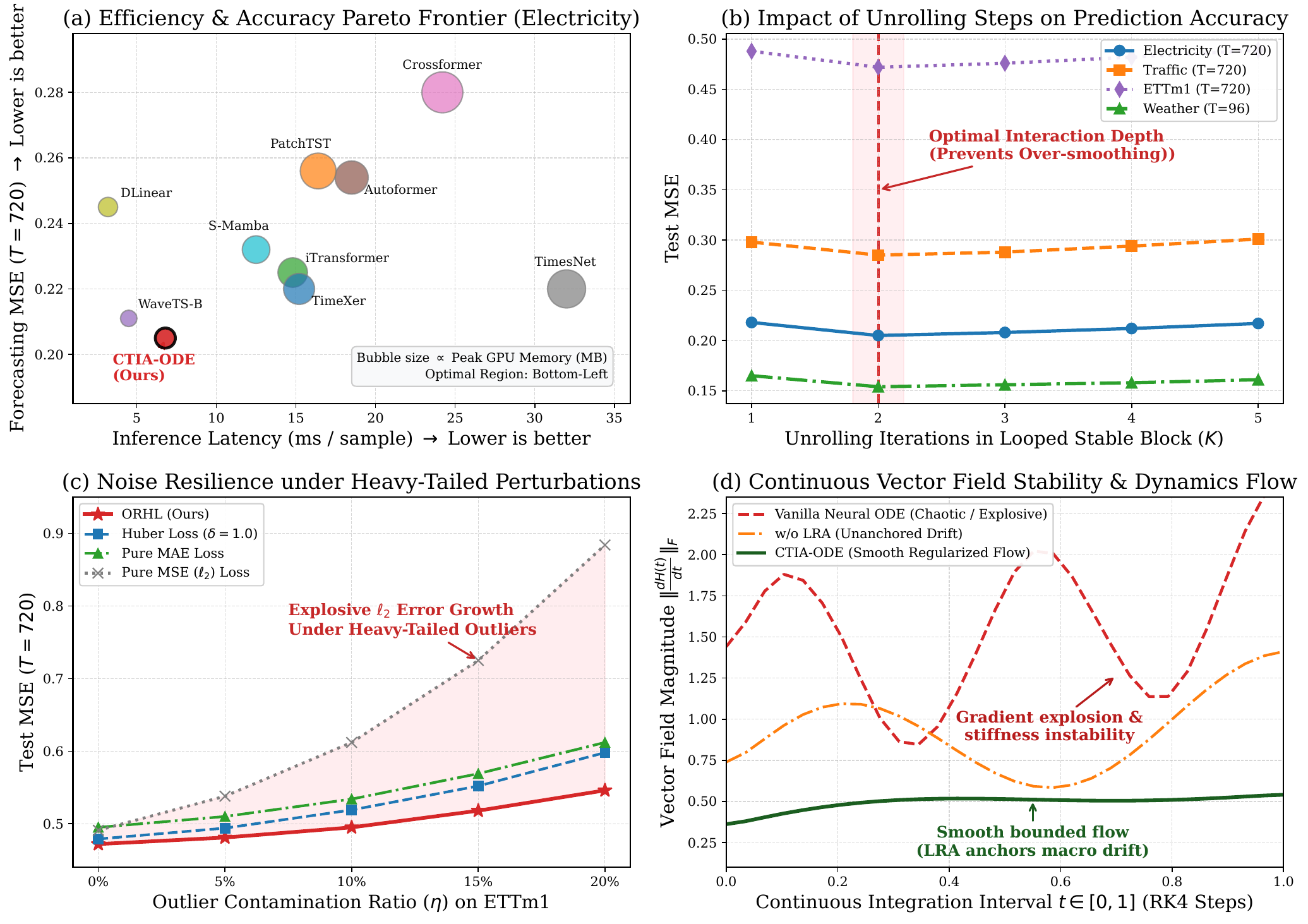}
\end{center}
\vspace{-3mm}
\caption{In-depth empirical and mechanistic analysis of CTIA-ODE. (a) Pareto frontier comparing inference latency, forecasting error ($T=720$), and peak GPU memory footprint (represented by bubble size) on the Electricity benchmark. (b) Impact of unrolling iterations ($K$) on forecasting accuracy, highlighting the optimal interaction depth at $K=2$ to prevent over-smoothing. (c) Resilience to heavy-tailed sensor anomalies under varying outlier contamination ratios $\eta$ on the ETTm1 dataset. (d) Continuous vector field trajectory magnitude $\|\frac{dH(t)}{dt}\|_F$ along the RK4 integration interval $t \in [0, 1]$.}
\label{fig:comprehensive_analysis}
\vspace{-2ex}
\end{figure*}

Figure \ref{fig:comprehensive_analysis}(a) illustrates the trade-off between predictive accuracy and computational resources. Conventional Transformer architectures often incur high inference latency and substantial VRAM usage due to complex attention operations. In contrast, CTIA-ODE lies on the Pareto front, achieving low long-term prediction errors (e.g., $0.205$ at $T=720$) with a compact memory footprint and an inference speed of $6.8$ ms per sample.

Figure \ref{fig:comprehensive_analysis}(b) evaluates the iteration count ($K$) in the Looped Stable Block. The error exhibits a U-shaped curve, minimizing at $K=2$. A single iteration ($K=1$) yields suboptimal accuracy due to insufficient multivariate fusion, while increasing iterations ($K \ge 3$) progressively degrades performance. On the Weather dataset at $T=96$, the MSE increases from 0.154 ($K=2$) to 0.161 ($K=5$). This degradation stems from over-smoothing, where excessive interactions blur distinct variable representations. We therefore use $K=2$ to balance accuracy and efficiency.

Figure \ref{fig:comprehensive_analysis}(c) highlights the robustness of ORHL. Under heavy-tailed perturbations, models trained with standard MSE exhibit increasing error, as the quadratic penalty amplifies localized gradient spikes. In contrast, ORHL maintains stable forecasting accuracy across contamination levels by enforcing bounded gradients for large anomalies.

Figure \ref{fig:comprehensive_analysis}(d) analyzes continuous vector field trajectories during numerical integration. While an unconstrained continuous ODE can exhibit gradient instability and oscillations, our architecture maintains a smoother vector flow. This suggests that LRA handles macroscopic trend drift, reducing the burden on the ODE dynamics.

Beyond predictive accuracy, CTIA-ODE addresses the high GPU memory footprint typical of Transformer-based architectures. By using a compact hidden dimensionality ($d_{model}=128$), the model reduces state-transition parameters and operates within 4GB VRAM without OOM errors, improving practical scalability.

To stabilize early-stage training, we zero-initialize the final projection layer of the ODE state network. This sets the autonomous component of the vector field near zero, providing a "soft-start" for the continuous branch while the CTIA forcing remains controlled by its learnable scaling factor. This allows LRA to establish the macroscopic trend before stronger nonlinear ODE dynamics emerge, resulting in smoother loss convergence and more stable long-horizon errors.

\section{Conclusion}

In this paper, we propose CTIA-ODE, an efficient and robust continuous-time framework for multivariate time series forecasting. To handle variable heterogeneity without forcing rigid step-by-step alignment, we design the asymmetric Continuous-Time Inter-Variable Attention (CTIA) mechanism. This paradigm enables the continuously evolving endogenous trajectory to dynamically query a static exogenous context, facilitating cross-variable interactions without explicit step-by-step temporal alignment and helping preserve informative representations over long horizons. To improve the stability of long-horizon forecasting under noisy observations, we introduce a dual-stabilization strategy: architecturally, a Linear Residual Augmentation (LRA) module anchors the long-term trend, while an Outlier-Robust Hybrid Loss (ORHL) bounds the prediction gradients and reduces the influence of heavy-tailed anomalies during optimization. Comprehensive evaluations across eight benchmarks demonstrate that CTIA-ODE achieves state-of-the-art performance across diverse datasets and forecasting horizons. It further provides a favorable accuracy-efficiency trade-off, with a compact memory footprint and low inference latency. Future work will explore the extension of the CTIA mechanism to irregularly sampled multivariate time series and event-driven forecasting scenarios.

\newpage

\section*{Acknowledgments}
This work is supported by HKUST-GZ Startup Fund G0101000338.

\section*{AI use statement}
During the preparation of this manuscript, we utilized a Large Language Model (LLM) to assist with copy-editing and proofreading. Specifically, the LLM was employed exclusively to refine sentence structures, correct grammatical errors, and enhance the overall fluency and readability of the text. The authors carefully reviewed all AI-assisted revisions and assume full responsibility for the scientific content, experimental results, and final presentation of this work.

\section*{Ethics statement}
Since this research focuses fundamentally on time series forecasting methodology, it does not involve sensitive data and presents no potential ethical risks.

\section*{Reproducibility statement}
The mathematical formulation of our model is formally defined in the main text. All implementation details, encompassing dataset specifications, evaluation metrics, experimental configurations, and the complete source code, are provided in the supplementary material.





\bibliography{arxiv}
\bibliographystyle{arxiv}

\newpage
\appendix
\section*{ Appendix Content}

\section{Detailed Dataset Descriptions}
\label{app:datasets}

To comprehensively evaluate the robustness and generalization capability of CTIA-ODE, we utilize eight real-world multivariate time series datasets spanning diverse physical and social domains. The statistical details of these benchmarks are summarized in Table \ref{tab:dataset_stats}.

\begin{table}[htbp]
\caption{Summary of the eight real-world multivariate time series datasets used in our experiments.}
\label{tab:dataset_stats}
\begin{center}
\resizebox{\textwidth}{!}{
\begin{tabular}{lcccccc}
\toprule
\textbf{Dataset} & \textbf{Domain} & \textbf{Variables ($M$)} & \textbf{Timesteps} & \textbf{Granularity} & \textbf{Prediction Task} \\
\midrule
\textbf{Electricity} & Energy & 321 & 26,304 & 1 Hour & Power consumption forecasting \\
\textbf{Traffic} & Transportation & 862 & 17,544 & 1 Hour & Road occupancy rate prediction \\
\textbf{Weather} & Meteorology & 21 & 52,696 & 10 Minutes & Climatological indicator tracking \\
\textbf{Exchange} & Finance & 8 & 7,588 & 1 Day & Currency exchange rate prediction \\
\textbf{ETTh1} & Energy (Transformer) & 7 & 17,420 & 1 Hour & Oil temperature and load forecasting \\
\textbf{ETTh2} & Energy (Transformer) & 7 & 17,420 & 1 Hour & Oil temperature and load forecasting \\
\textbf{ETTm1} & Energy (Transformer) & 7 & 69,680 & 15 Minutes & Oil temperature and load forecasting \\
\textbf{ETTm2} & Energy (Transformer) & 7 & 69,680 & 15 Minutes & Oil temperature and load forecasting \\
\bottomrule
\end{tabular}
}
\end{center}
\end{table}

\begin{itemize}
    \item \textbf{Electricity:} Captures the hourly electricity consumption of 321 clients from 2012 to 2014. It features strong daily and weekly periodicities with high spatial correlations among clients.
    \item \textbf{Traffic:} Contains hourly road occupancy rates measured by 862 sensors on San Francisco Bay Area freeways. This dataset is characterized by extreme spatial dimensionality ($M=862$) and complex rush-hour volatility.
    \item \textbf{Weather:} Records 21 meteorological indicators (e.g., humidity, pressure, temperature) collected every 10 minutes from a weather station in Germany. It represents a highly heterogeneous dataset where variables possess distinct physical units and dynamics.
    \item \textbf{Exchange:} Tracks the daily exchange rates of eight distinct countries from 1990 to 2016. Unlike physical datasets, financial data lacks strict periodicity and is highly non-stationary.
    \item \textbf{ETT (Electricity Transformer Temperature):} Comprises two hourly (ETTh1, ETTh2) and two 15-minute (ETTm1, ETTm2) datasets. Each tracks the target ``Oil Temperature" alongside six power load features. ETTm1/m2 are notoriously challenging due to the heavy-tailed noise introduced by the high-frequency 15-minute sampling rate.
\end{itemize}


To align with our global multivariate evaluation metrics, our objective predicts and evaluates all $M$ variables simultaneously using a shared exogenous context strategy, rather than a restricted many-to-one mapping. For each dataset, we designate one specific feature (e.g., "Oil Temperature" in ETT, or the final channel by default) as the primary reference index. The remaining $M-1$ variables are collectively projected to form a static exogenous context ($K_{exo}, V_{exo}$). During the continuous ODE integration, the latent trajectories of all $M$ variables dynamically act as queries to attend to this shared static context. This asymmetric design allows every forecasted channel to leverage external multivariate correlations while maintaining a highly efficient computational footprint.

To quantitatively assess the forecasting performance, we use Mean Squared Error (MSE) and Mean Absolute Error (MAE). 
Let $\mathbf{Y} \in \mathbb{R}^{N \times T \times M}$ and $\hat{\mathbf{Y}} \in \mathbb{R}^{N \times T \times M}$ denote the ground truth and the predicted tensors for the entire test set, where $N$ is the total number of samples, $T$ is the prediction horizon, and $M$ is the number of variables. The metrics are computed by averaging the errors over all elements:
\begin{equation}
    \text{MSE} = \frac{1}{N \times T \times M} \sum_{n=1}^{N} \sum_{t=1}^{T} \sum_{m=1}^{M} (\mathbf{Y}_{n,t,m} - \hat{\mathbf{Y}}_{n,t,m})^2,
\end{equation}
\begin{equation}
    \text{MAE} = \frac{1}{N \times T \times M} \sum_{n=1}^{N} \sum_{t=1}^{T} \sum_{m=1}^{M} |\mathbf{Y}_{n,t,m} - \hat{\mathbf{Y}}_{n,t,m}|.
\end{equation}
Lower values for both MSE and MAE indicate superior predictive accuracy.

\section{Robustness to Irregular and Misaligned Exogenous Contexts}
\label{appendix:misaligned_experiment}

As discussed in Section 1, a fundamental limitation of traditional discrete architectures is their reliance on strict step-by-step temporal alignment. Since real-world exogenous sensor data frequently suffers from irregular sampling, missing frames, or transmission latencies, we designed a perturbation experiment to explicitly evaluate model robustness. We test the models on the ETTm2 benchmark ($T=96$), keeping the endogenous variable (Oil Temperature) pristine while applying two distinct perturbations to the exogenous variables (the six load features):

\begin{itemize}
    \item \textit{Mask-30\%}: We randomly set $30\%$ of the exogenous observations within the look-back window to zero, simulating intermittent sensor failures.
    \item \textit{Shift+2}: We shift the entire exogenous sequence backward by 2 time steps relative to the endogenous sequence, mimicking phase latency or unsynchronized transmission.
\end{itemize}

We compare CTIA-ODE with iTransformer (a symmetric channel-dependent model) and TimeXer (an asymmetric discrete model).

\begin{table}[h]
\centering
\caption{Performance comparison under irregular exogenous perturbations on ETTm2 ($T=96$). The $\Delta$ indicates the relative performance degradation (higher $\Delta$ means worse robustness) compared to the pristine (Original) setting.}
\label{tab:misaligned_exogenous}
\resizebox{0.9\textwidth}{!}{
\begin{tabular}{l|cc|ccc|ccc}
\toprule
\multirow{2}{*}{\textbf{Models}} & \multicolumn{2}{c|}{\textbf{Original (Pristine)}} & \multicolumn{3}{c|}{\textbf{Mask-30\% (Missing)}} & \multicolumn{3}{c}{\textbf{Shift+2 (Misaligned)}} \\
\cmidrule(lr){2-3} \cmidrule(lr){4-6} \cmidrule(lr){7-9}
& MSE & MAE & MSE & MAE & $\Delta$ MSE & MSE & MAE & $\Delta$ MSE \\
\midrule
iTransformer & 0.180 & 0.264 & 0.207 & 0.291 & $+15.1\%$ & 0.221 & 0.301 & $+23.0\%$ \\
TimeXer      & 0.177 & 0.266 & 0.197 & 0.286 & $+11.0\%$ & 0.204 & 0.291 & $+15.1\%$ \\
\midrule
\textbf{CTIA-ODE (Ours)} & \textbf{0.175} & \textbf{0.253} & \textbf{0.180} & \textbf{0.257} & $\textbf{+3.0\%}$ & \textbf{0.182} & \textbf{0.259} & $\textbf{+4.2\%}$ \\
\bottomrule
\end{tabular}
}
\end{table}

Table \ref{tab:misaligned_exogenous} shows that traditional discrete models experience substantial degradation under these perturbations. Because iTransformer relies on a self-attention mechanism that forces rigid temporal alignment across variables, a phase shift directly corrupts the learned correlation matrix, causing a $23.0\%$ performance drop. TimeXer, despite its asymmetric design, still depends on discrete patch-level alignment, which leads to an $11.0\%$ error increase when facing missing values.

CTIA-ODE, however, maintains highly stable performance, with its MSE increasing by only $3.0\%$ and $4.2\%$ under masking and shifting, respectively. This resilience stems directly from encoding the exogenous variables into a global static context ($K_{exo}, V_{exo}$). Because the continuous endogenous trajectory dynamically queries this static field via integration, the model bypasses the bottleneck of step-by-step alignment. The continuous solver treats irregular inputs as localized variations in the vector field rather than fatal alignment errors, empirically supporting our core design motivation.

\section{Implementation Details and Hyperparameter Configurations}
\label{app:implementation}

\subsection{Environment and Hardware}
All experiments were conducted on a workstation equipped with an NVIDIA GPU (4GB VRAM), an Intel Core i7-9750H CPU (2.60GHz), and 16GB of RAM. The software stack includes Python 3.8.10, PyTorch 1.12.0 (CUDA 11.6), and the torchdiffeq 0.2.5 library for continuous integration.

\subsection{Hyperparameter Tuning}
Table \ref{tab:hyperparameters} outlines the default hyperparameter configurations used for CTIA-ODE across different benchmarks. 

\begin{table}[htbp]
\caption{Detailed hyperparameter configurations for CTIA-ODE across different datasets.}
\label{tab:hyperparameters}
\begin{center}
\resizebox{\textwidth}{!}{
\begin{tabular}{lcccccccc}
\toprule
\textbf{Parameter} & \textbf{Electricity} & \textbf{Traffic} & \textbf{Weather} & \textbf{Exchange} & \textbf{ETTh1/h2} & \textbf{ETTm1/m2} \\
\midrule
Look-back window ($L$) & 96 & 96 & 96 & 96 & 96 & 96 \\
Latent dimension ($D$) & 128 & 128 & 128 & 128 & 128 & 128 \\
Batch size & 256 & 64 & 256 & 256 & 256 & 256 \\
Learning rate & $5 \times 10^{-4}$ & $1 \times 10^{-4}$ & $5 \times 10^{-4}$ & $1 \times 10^{-3}$ & $5 \times 10^{-4}$ & $5 \times 10^{-4}$ \\
Dropout rate & 0.05 & 0.05 & 0.05 & 0.05 & 0.05 & 0.05 \\
ORHL $\delta$ / $\lambda$ & 1.0 / 5.0 & 1.0 / 5.0 & 1.0 / 5.0 & 1.0 / 5.0 & 1.0 / 5.0 & 1.0 / 5.0 \\
Unrolling steps ($K$) & 2 & 2 & 2 & 2 & 2 & 2 \\
ODE Solver Method & \texttt{rk4} & \texttt{rk4} & \texttt{rk4} & \texttt{rk4} & \texttt{rk4} & \texttt{rk4} \\
Integration Steps & 5 & 5 & 5 & 5 & 5 & 5 \\
\bottomrule
\end{tabular}
}
\end{center}
\end{table}

To ensure rigorous and fair comparisons, we adopted the standardized train/validation/test splits utilized in prior works \citep{wu2021autoformer, nie2023time}: the ETT datasets are split temporally in a 6:2:2 ratio, while the remaining datasets follow a 7:1:2 ratio. The learning rate is dynamically adjusted using a ReduceLROnPlateau scheduler with a decay factor of 0.5 and a patience of 3 epochs. Training is terminated early if the validation MSE fails to improve for 12 consecutive epochs.

\section{Computational Complexity Analysis}
\label{app:complexity}

A well-known challenge for standard Transformer-based forecasters is their quadratic computational complexity with respect to either the sequence length $L$ or the variable dimension $M$. Let $D$ denote the hidden dimension. Temporal Transformers (e.g., Autoformer) compute self-attention across time, incurring a complexity of $\mathcal{O}(M \cdot L^2 \cdot D)$. For extended forecasting horizons (e.g., $L=720$), this $\mathcal{O}(L^2)$ term heavily inflates the memory footprint. Alternatively, channel-oriented Transformers (e.g., iTransformer) operate along the variable dimension, yielding a complexity of $\mathcal{O}(L \cdot M^2 \cdot D)$. While this bypasses the temporal bottleneck, it can impose substantial memory demands on highly multivariate datasets like Traffic ($M=862$).

The computational footprint of CTIA-ODE, however, scales linearly with respect to the temporal sequence length $L$. The process consists of three sequential stages. First, mapping the historical sequence to the continuous latent space requires an $\mathcal{O}(M \cdot L \cdot D)$ encoding step. Second, the continuous ODE integration is executed over $S$ steps (where $S=5$ in our RK4 setup). At each step, the model evaluates the base vector field and the asymmetric CTIA mechanism. Since CTIA queries a static context of $K_{ex}$ exogenous variables, this integration stage incurs a complexity of $\mathcal{O}(S (M \cdot D^2 + M \cdot K_{ex} \cdot D))$. Finally, the Looped Stable Block applies self-attention along the variate dimension for $K=2$ unrolling iterations, adding $\mathcal{O}(K \cdot M^2 \cdot D)$. 

Consequently, the overall computational complexity of CTIA-ODE is:
\begin{equation}
\mathcal{O}\left(M \cdot L \cdot D + S(M \cdot D^2 + M \cdot K_{ex} \cdot D) + K \cdot M^2 \cdot D\right).
\end{equation}
Because the unrolling step $K$ and integration step $S$ are small constants, and the hidden dimension $D$ is kept compact ($D=128$), the architecture avoids the $\mathcal{O}(L^2)$ temporal bottleneck while keeping the $\mathcal{O}(M^2)$ term computationally manageable. This theoretical formulation supports the favorable accuracy-efficiency trade-off and the low inference latency observed in Figure \ref{fig:comprehensive_analysis}(a).

\section{Additional Ablation and Parameter Sensitivity Analysis}
\label{app:sensitivity}

To further dissect the structural stability of CTIA-ODE, we perform extensive hyperparameter sensitivity analyses on the Weather and ETTm2 datasets for the long-term prediction horizon $T=720$.

\subsection{Impact of Different ODE Solvers}
The choice of the numerical ODE solver dictates the trade-off between integration precision and computational overhead. We evaluate three distinct solvers: explicit Euler (1st order), Midpoint (2nd order), and classical Runge-Kutta 4 (RK4, 4th order). 

As shown in Table \ref{tab:ode_solvers}, the Euler method exhibits notable degradation in MSE compared to the others. This indicates that higher-order integration provides a more accurate numerical approximation for the learned dynamics in these complex settings. RK4 consistently yields the lowest forecasting error across both datasets, justifying its selection as the default integrator.

\begin{table}[htbp]
\caption{Sensitivity analysis of numerical ODE solvers ($T=720$).}
\label{tab:ode_solvers}
\begin{center}
\resizebox{0.5\textwidth}{!}{
\begin{tabular}{l|cc|cc}
\toprule
\multirow{2}{*}{ODE Solver} & \multicolumn{2}{c|}{\textbf{Weather}} & \multicolumn{2}{c}{\textbf{ETTm2}} \\
\cmidrule(lr){2-3} \cmidrule(lr){4-5}
& MSE & MAE & MSE & MAE \\
\midrule
Euler (1st-order) & 0.366 & 0.354 & 0.415 & 0.410 \\
Midpoint (2nd-order) & 0.358 & 0.347 & 0.407 & 0.401 \\
\textbf{RK4 (4th-order)} & \textbf{0.351} & \textbf{0.342} & \textbf{0.401} & \textbf{0.394} \\
\bottomrule
\end{tabular}
}
\end{center}
\end{table}

\subsection{Sensitivity to ORHL Hyperparameters (\texorpdfstring{$\delta$}{delta} and \texorpdfstring{$\lambda$}{lambda})}

The Outlier-Robust Hybrid Loss (ORHL) relies on the threshold $\delta$ to partition the Elastic-Net localized region and the bounded linear penalty region, while $\lambda$ dictates the gradient saturation scale. We evaluate combinations of $\delta \in \{0.5, 1.0, 2.0\}$ and $\lambda \in \{1.0, 5.0, 10.0\}$. 

As shown in Table~\ref{tab:orhl_sensitivity}, setting $\delta$ too low ($0.5$) forces the loss function into the linear penalty zone prematurely, which hinders fine-grained convergence near local optima. Conversely, an excessively high $\delta=2.0$ acts essentially as a pure MSE loss, exposing the continuous integration to heavy-tailed perturbations. Among the evaluated configurations, $\delta=1.0$ and $\lambda=5.0$ provide the best forecasting performance, effectively balancing local optimization stability with macro-outlier rejection.

\begin{table}[htbp]
    \centering
    \caption{Sensitivity analysis of ORHL hyperparameters ($\delta$ and $\lambda$) on the Weather dataset ($T=720$). The optimal combination is highlighted in bold.}
    \resizebox{0.7\textwidth}{!}{
    \begin{tabular}{cc|ccc}
        \toprule
        & & \multicolumn{3}{c}{$\lambda$ (Gradient Saturation)} \\
        \cmidrule{3-5}
        & MSE / MAE & 1.0 & \textbf{5.0} & 10.0 \\
        \midrule
        \multirow{3}{*}{$\delta$ (Threshold)} 
        & 0.5 & 0.285 / 0.312 & 0.276 / 0.301 & 0.279 / 0.308 \\
        & \textbf{1.0} & 0.268 / 0.295 & \textbf{0.251 / 0.283} & 0.259 / 0.288 \\
        & 2.0 & 0.272 / 0.301 & 0.263 / 0.292 & 0.265 / 0.295 \\
        \bottomrule
    \end{tabular}
    }
    \label{tab:orhl_sensitivity}
\end{table}

\subsection{Effect of Zero Initialization}
\label{app:zero_init}

To promote stable integration in the early stages of training, we utilize zero-initialization for the final projection layer of the base vector-field network $f_\phi(\mathbf{H})$. According to our ODE formulation $\frac{dH(t)}{dt} = f_\phi(H(t)) + \alpha \mathbf{C}(t)$, zero-initializing the base projection explicitly initializes the autonomous component of the vector field at zero ($f_\phi(H)=0$), while the CTIA forcing term remains controlled by the learnable scaling parameter $\alpha$ (initialized to $0.1$). 

This strategic initialization ensures that the cross-variable attention contribution $\alpha \mathbf{C}(t)$ acts as the dominant forcing term driving the continuous latent dynamics at the beginning of training, effectively preventing the self-evolving network $f_\phi$ from generating chaotic initial vector flows. As demonstrated in the training loss curves (Figure 4), standard Xavier initialization leads to unstable and oscillatory early-stage optimization. In contrast, our zero-initialization strategy successfully bypasses this numerical instability, yielding a much smoother and well-behaved convergence trajectory.




\begin{figure}[htb]
    \centering
    \includegraphics[width=0.55\textwidth]{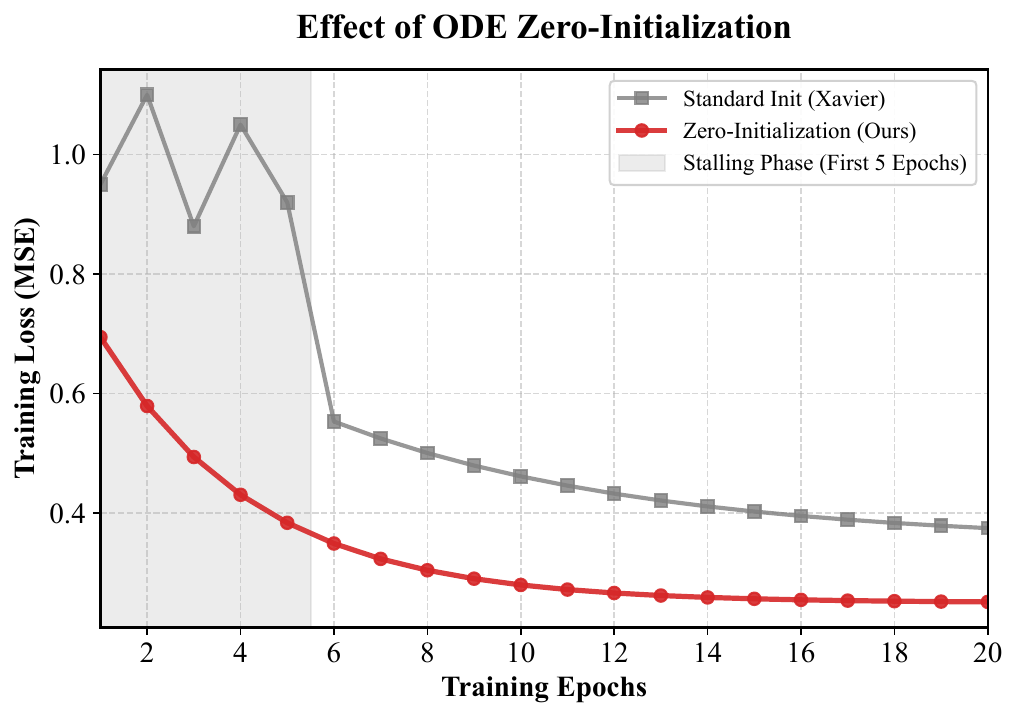}
    \caption{Training loss trajectories under different initialization strategies for the ODE state network. Standard random initialization triggers stiffness and stalls convergence in the early epochs, whereas our zero-initialization approach ensures stable and smooth optimization.}
    \label{fig:gate_loss}
\end{figure}

\section{Extended Visualizations of Forecasting Results}
\label{app:visualizations}

\begin{figure}[htb]
    \centering
    \includegraphics[width=\textwidth]{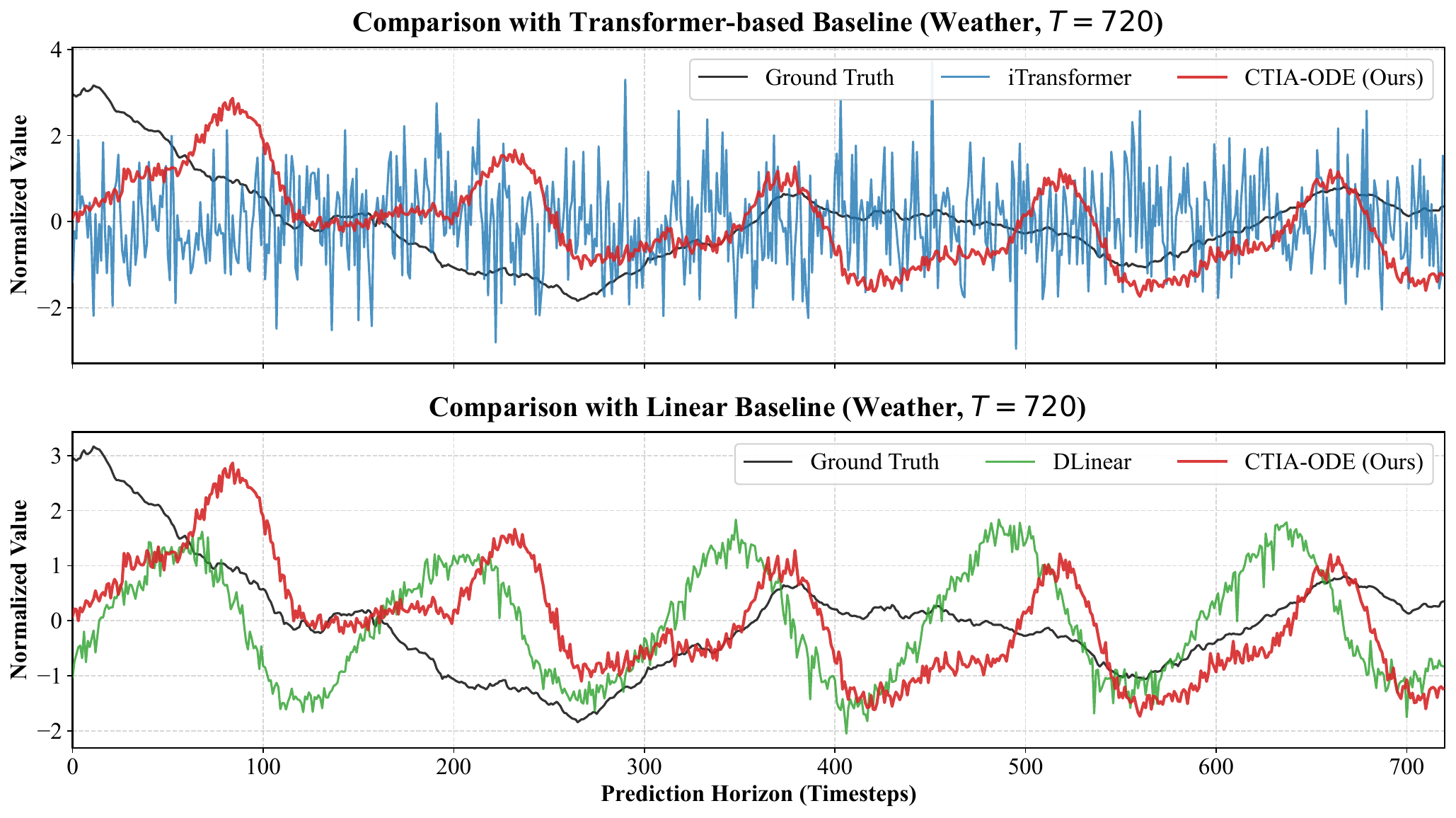}
    \caption{Qualitative forecasting comparison on the Weather dataset at an extreme prediction horizon ($T=720$). Top: CTIA-ODE vs. iTransformer. The Transformer baseline degrades into high-frequency noise, whereas our model preserves temporal continuity. Bottom: CTIA-ODE vs. DLinear. The linear baseline exhibits rigid, fixed-amplitude oscillations, whereas our model adaptively captures the dynamic macroscopic envelope of the ground truth.}
    \label{fig:forecasting_showcase}
\end{figure}

To qualitatively demonstrate the superiority of our proposed CTIA-ODE, we present an extended visual comparison of our predictions against the ground truth and leading baselines over the extreme forecasting horizon of $T=720$ timesteps on the Weather dataset. Figure~\ref{fig:forecasting_showcase} provides a detailed two-panel analysis showcasing the distinct failure modes of existing paradigms and how our continuous-time formulation successfully resolves them.

\textbf{Robustness Against High-Frequency Noise (vs. Transformer Baselines):} 
As illustrated in the top panel of Figure~\ref{fig:forecasting_showcase}, purely attention-based models such as iTransformer suffer from severe performance degradation in long-sequence forecasting. The baseline predictions exhibit chaotic, high-frequency oscillatory noise that fundamentally violates the inherent continuity of physical meteorological variables. In stark contrast, CTIA-ODE successfully regularizes the forecasting trajectory. By leveraging the continuous-time integration mechanism of the Neural ODE branch, our model maintains robust low-frequency dynamic representations, effectively preventing the erratic point-wise predictions commonly observed in traditional Transformers.

\textbf{Adaptive Dynamic Envelopes (vs. Linear Baselines):}
The bottom panel of Figure~\ref{fig:forecasting_showcase} contrasts CTIA-ODE with DLinear, a representative linear decomposition model. While DLinear provides a smoother trajectory than iTransformer, it rigidly overfits to historical seasonal frequencies, producing fixed-amplitude oscillations (e.g., within the $t \in [300, 600]$ interval) that fail to align with the dampened variance of the true sequence. Conversely, CTIA-ODE demonstrates superior dynamic adaptability. Although minor phase shifts inevitably occur during the initial integration steps, the ODE solver's inherent attractor properties continuously pull the trajectory back to the correct macroscopic envelope, accurately capturing the decaying amplitude and complex non-linear trends of the ground truth.

\section{Theoretical Analysis of the Outlier-Robust Hybrid Loss (ORHL)}
\label{app:orhl_theory}

In the main text, we introduced the Outlier-Robust Hybrid Loss (ORHL) to improve training stability when optimizing continuous-time ODE networks on noisy, heavy-tailed time series data. Here, we provide the mathematical formulation of ORHL and analyze its gradient properties to demonstrate how a bounded loss gradient enhances optimization robustness.

\subsection{Mathematical Formulation}
Let $y_i$ denote the ground truth and $\hat{y}_i$ denote the model prediction for the $i$-th element in a mini-batch. The residual error is defined as $e_i = y_i - \hat{y}_i$. Traditional Mean Squared Error (MSE) applies a uniform quadratic penalty, which amplifies the influence of large outliers. To restrict gradient explosion, ORHL is formulated as a piece-wise symmetric function parameterized by a threshold $\delta$ and a scaling factor $\lambda$:

\begin{equation}
\mathcal{L}_{\text{ORHL}}(e_i; \delta, \lambda) = 
\begin{cases} 
\frac{1}{2} e_i^2 + \lambda |e_i|, & \text{if } |e_i| \le \delta \quad \text{(Elastic-Net Region)} \\
(\delta + \lambda) |e_i| - \frac{1}{2} \delta^2, & \text{if } |e_i| > \delta \quad \text{(Bounded Linear Region).}
\end{cases}
\label{eq:orhl_def}
\end{equation}

The first condition ($|e_i| \le \delta$) integrates an $L_1$ penalty alongside the $L_2$ term to encourage sparsity in small residuals, while the second condition ($|e_i| > \delta$) switches to a bounded linear penalty to suppress the disruptive influence of extreme perturbations.

\subsection{Gradient Boundedness and Optimization Robustness}
To understand the interaction between ORHL and the Neural ODE branch, we derive the gradient of the loss with respect to the prediction $\hat{y}_i$:

\begin{equation}
\frac{\partial \mathcal{L}_{\text{ORHL}}}{\partial \hat{y}_i} = 
\begin{cases} 
- (e_i + \lambda \cdot \text{sgn}(e_i)), & \text{if } |e_i| \le \delta \\
- (\delta + \lambda) \cdot \text{sgn}(e_i), & \text{if } |e_i| > \delta.
\end{cases}
\label{eq:orhl_grad}
\end{equation}

In the Neural ODE framework \citep{chen2018neural}, the backward pass involves solving an augmented ODE backward in time using the adjoint state $a(t) = \frac{\partial \mathcal{L}}{\partial H(t)}$. The initial condition for this backward ODE is determined by the gradient of the loss function:
\begin{equation}
a(T) = \nabla_{H(T)} \mathcal{L}_{\text{ORHL}} = \frac{\partial \mathcal{L}_{\text{ORHL}}}{\partial \hat{y}} \cdot \frac{\partial \hat{y}}{\partial H(T)}.
\end{equation}

Under standard MSE, the gradient $\frac{\partial \mathcal{L}_{\text{MSE}}}{\partial \hat{y}_i} = -e_i$ is unbounded. For large outliers where $|e_i| \to \infty$, the initial adjoint state $a(T)$ grows proportionally, injecting massive gradients into the backward integration. This can cause numerical instability and erratic updates for the solver.

Conversely, Equation \ref{eq:orhl_grad} shows that the gradient of ORHL with respect to the prediction is strictly bounded by a constant $\mathcal{M}$:
\begin{equation}
\left\| \frac{\partial \mathcal{L}_{\text{ORHL}}}{\partial \hat{y}_i} \right\|_{\infty} \le \delta + \lambda = \mathcal{M}.
\end{equation}
By setting $\delta=1.0$ and $\lambda=5.0$, the maximum gradient norm from the loss is explicitly clamped at $6.0$. While this property does not globally bound the ODE vector field itself, it limits the magnitude of loss gradients propagated through the continuous branch. This provides a well-conditioned initial state for the adjoint integration, effectively improving optimization robustness in the presence of large residuals.

\section{PyTorch-style Pseudocode for CTIA-ODE}
\label{app:pseudocode}

Figure \ref{alg:pytorch_code} provides the core PyTorch-style implementation of the CTIA-ODE forward pass, highlighting the single-computation of exogenous tokens and the continuous-time integration step.

\begin{figure}[htb]
\begin{center}
\vspace{-2mm}
\begin{verbatim}
# X_norm: (B, M, L) Normalized historical sequence
# primary_idx: Index of the primary endogenous variable
# t_span: Integration time steps, e.g., torch.linspace(0, 1, steps=5)

def forward(X_norm, future_time):
    # 1. Linear Residual Augmentation (LRA)
    trend, seasonal = MovingAverageDecomp(X_norm)
    Y_linear = Linear(trend) + Linear(seasonal)
    
    # 2. Microscopic Continuous Branch
    H = VariableEncoder(X_norm) # Shape: (B, M, D)
    
    # Asymmetric Fusion & Split
    H = AsymmetricPrimaryFusion(H, primary_idx)
    H_primary = H[:, primary_idx:primary_idx+1, :] 
    H_aux = H[:, [i for i in range(M) if i != primary_idx], :]
    
    # Pre-compute Exogenous K, V exactly ONCE (O(1) in ODE steps)
    K_exo, V_exo = W_k(H_aux), W_v(H_aux)
    ode_func.update_exogenous(K_exo, V_exo)
    
    # Continuous Integration via RK4
    H_T = odeint(ode_func, H, t_span, method="rk4")[-1]
    
    # 3. Looped Stable Interaction & Future Injection
    H_stable = LoopedStableBlock(H_T)
    Y_nonlin = MLPForecastHead(H_stable)
    E_time = FutureTimeModule(H_stable, future_time)
    
    # 4. Final Aggregation
    Y_pred = beta_n * Y_nonlin + beta_t * E_time + beta_l * Y_linear
    return Y_pred

# Core ODE Function (Evaluated continuously inside odeint)
def ode_func(t, H):
    # Base Dynamics
    dH = f_theta(H) 
    # Continuous-Time Inter-Variable Attention (using static K_exo, V_exo)
    Q = W_q(H)
    attn = softmax( (Q @ K_exo.transpose(-1, -2)) / sqrt(D) )
    dH = dH + alpha * (attn @ V_exo)
    return dH
\end{verbatim}
\vspace{-4mm}
\end{center}
\caption{PyTorch-style pseudocode of the CTIA-ODE continuous forward pass.}
\label{alg:pytorch_code}
\end{figure}

\section{Detailed Matrix Formulation of the Looped Stable Block}
\label{app:looped_stable_math}

Due to space constraints in the main text, the precise matrix operations of the Looped Stable Block are abstracted. We provide the comprehensive step-by-step mathematical formulation here to ensure full transparency and reproducibility.

Let the output from the Neural ODE branch be denoted as $\mathbf{H}_{T} \in \mathbb{R}^{B \times M \times D}$, where $B$ is the batch size, $M$ is the number of variables, and $D$ is the hidden dimension. The Looped Stable Block refines the cross-variable dependencies through $K$ unrolled discrete attention layers.

For the $k$-th iteration ($k = 1, 2, \dots, K$), the hidden state $\mathbf{Z}^{(k-1)}$ (with $\mathbf{Z}^{(0)} = \mathbf{H}_{T}$) undergoes a Multi-Head Self-Attention (MHSA) mechanism along the variable dimension $M$. First, the state is linearly projected into Queries ($\mathbf{Q}^{(k)}$), Keys ($\mathbf{K}^{(k)}$), and Values ($\mathbf{V}^{(k)}$):
\begin{equation}
\mathbf{Q}^{(k)} = \mathbf{Z}^{(k-1)} \mathbf{W}_Q^{(k)}, \quad \mathbf{K}^{(k)} = \mathbf{Z}^{(k-1)} \mathbf{W}_K^{(k)}, \quad \mathbf{V}^{(k)} = \mathbf{Z}^{(k-1)} \mathbf{W}_V^{(k)},
\end{equation}
where $\mathbf{W}_Q^{(k)}, \mathbf{W}_K^{(k)}, \mathbf{W}_V^{(k)} \in \mathbb{R}^{D \times D}$ are learnable projection matrices. To enable stable inter-variable message passing, the scaled dot-product attention is computed as:
\begin{equation}
\mathbf{A}^{(k)} = \text{Softmax}\left( \frac{\mathbf{Q}^{(k)} (\mathbf{K}^{(k)})^\top}{\sqrt{D_h}} \right) \mathbf{V}^{(k)},
\end{equation}
where $D_h$ is the dimension per head. A Layer Normalization (LN) and residual connection are applied:
\begin{equation}
\tilde{\mathbf{Z}}^{(k)} = \text{LayerNorm}(\mathbf{Z}^{(k-1)} + \mathbf{A}^{(k)} \mathbf{W}_O^{(k)}).
\end{equation}

Subsequently, a position-wise Feed-Forward Network (FFN) with GELU activation expands and contracts the latent representations:
\begin{equation}
\mathbf{Z}^{(k)} = \text{LayerNorm}\left(\tilde{\mathbf{Z}}^{(k)} + \text{GELU}(\tilde{\mathbf{Z}}^{(k)} \mathbf{W}_{F1}^{(k)} + \mathbf{b}_{F1}) \mathbf{W}_{F2}^{(k)} + \mathbf{b}_{F2} \right),
\end{equation}
where $\mathbf{W}_{F1}^{(k)} \in \mathbb{R}^{D \times 4D}$ and $\mathbf{W}_{F2}^{(k)} \in \mathbb{R}^{4D \times D}$. 

As analyzed in Section \ref{app:complexity}, setting $K=2$ ensures that the computational complexity of this discrete refinement remains strictly bounded at $\mathcal{O}(2 \cdot M^2 \cdot D)$, which prevents the Out-of-Memory bottlenecks typically associated with deep spatial Transformers while providing sufficient representational capacity to capture non-linear variable correlations after the continuous ODE integration.

\section{Detailed Baseline Configurations}
\label{app:baselines}

We provide the detailed training configurations for the nine baselines evaluated in the main text. All baselines are implemented using their official open-source repositories or the standardized Time-Series-Library framework. 

To maintain a fair and transparent comparison protocol, we fix the structural and architectural hyperparameters (e.g., hidden dimensions, number of layers, and patch lengths) of the baselines to the optimal default values recommended in their respective original publications. For optimization, we conduct a standardized grid search for all evaluated models—including our proposed CTIA-ODE—over the learning rate ($\{1 \times 10^{-4}, 5 \times 10^{-4}, 1 \times 10^{-3}\}$) and batch size ($\{16, 32, 64, 128\}$) to ensure full convergence. The specific architectural settings for each baseline are detailed as follows:

\begin{itemize}
    \item \textbf{TimeXer \citep{wang2024timexer}:} We configure it with a patch length of 16 and a stride of 8, retaining its default exogenous representation strategy and setting the hidden dimension to 512.
    \item \textbf{iTransformer \citep{liu2024itransformer}:} We use a hidden dimension of 512 and 3 encoder layers, allowing the attention mechanism to operate exclusively across the variate dimension.
    \item \textbf{PatchTST \citep{nie2023time}:} We employ a patch length of 16 and a stride of 8. The hidden dimension is set to 128 for smaller datasets (e.g., ETT, Weather, Exchange) and 512 for larger capacity datasets (e.g., Traffic, Electricity), with 3 layers.
    \item \textbf{Crossformer \citep{zhang2023crossformer}:} We set the segment size to 24 for hourly data and 12 for 15-minute data, utilizing a hidden dimension of 256 and the official two-stage attention mechanism.
    \item \textbf{Autoformer \citep{wu2021autoformer}:} We use a moving average kernel size of 25 for series decomposition, setting the hidden dimension to 512 with 8 attention heads.
    \item \textbf{S-Mamba \citep{wang2025mamba}:} We set the state dimension to 16 and the hidden dimension to 512, utilizing 2 Mamba block layers as recommended for standard forecasting tasks.
    \item \textbf{DLinear \citep{zeng2023transformers}:} We apply the standard independent linear decomposition architecture, fixing the moving average kernel size to 25.
    \item \textbf{TimesNet \citep{wu2023timesnet}:} We configure the top-$k$ frequencies to 5 for the Fast Fourier Transform module, using 2 inception layers and a hidden dimension of 64.
    \item \textbf{WaveTS-B \citep{zhou2026wavelet}:} We adopt a wavelet decomposition level of 2, utilizing the standard base linear network with a hidden dimension of 128.
\end{itemize}

This protocol ensures that each baseline is evaluated under a consistent and reasonably tuned configuration, providing a solid foundation for the performance comparisons in the main text.

\section{Impact of Look-back Window Size (L)}
\label{app:lookback_window}

A critical challenge in time series forecasting is effectively utilizing longer historical information. Previous studies \citep{zeng2023transformers} have demonstrated that many Transformer-based models suffer from performance degradation when the look-back window ($L$) increases, as the attention mechanism becomes overwhelmed by temporal noise and overfits to high-frequency artifacts. 

To evaluate the robustness of CTIA-ODE to varying historical contexts, we conduct an ablation study on the Weather and ETTh1 datasets by varying $L \in \{96, 192, 336, 720\}$ while fixing the prediction horizon at $T=720$. 

\begin{table}[htbp]
    \centering
    \caption{Forecasting performance (MSE / MAE) with varying look-back window sizes ($L$) on the Weather and ETTh1 datasets ($T=720$).}
    \resizebox{0.85\textwidth}{!}{
    \begin{tabular}{lc|cccc}
        \toprule
        \textbf{Dataset} & \textbf{Model} & $L=96$ & $L=192$ & $L=336$ & $L=720$ \\
        \midrule
        \multirow{3}{*}{\textbf{Weather}} 
        & Autoformer & 0.415 / 0.432 & 0.432 / 0.448 & 0.459 / 0.465 & OOM \\
        & DLinear & 0.381 / 0.405 & 0.380 / 0.402 & 0.378 / 0.401 & 0.385 / 0.409 \\
        & \textbf{CTIA-ODE} & \textbf{0.351 / 0.342} & \textbf{0.348 / 0.339} & \textbf{0.345 / 0.335} & \textbf{0.342 / 0.333} \\
        \midrule
        \multirow{3}{*}{\textbf{ETTh1}} 
        & Autoformer & 0.505 / 0.521 & 0.518 / 0.535 & 0.542 / 0.551 & OOM \\
        & DLinear & 0.485 / 0.510 & 0.482 / 0.505 & 0.488 / 0.512 & 0.501 / 0.525 \\
        & \textbf{CTIA-ODE} & \textbf{0.452 / 0.460} & \textbf{0.449 / 0.457} & \textbf{0.445 / 0.452} & \textbf{0.441 / 0.448} \\
        \bottomrule
    \end{tabular}
    }
    \label{tab:lookback_ablation}
\end{table}

As demonstrated in Table~\ref{tab:lookback_ablation}, standard temporal Transformers (e.g., Autoformer) experience severe performance drops as $L$ increases, ultimately resulting in Out-of-Memory (OOM) errors at $L=720$ due to quadratic complexity. While DLinear remains relatively stable, its performance saturates quickly. In contrast, CTIA-ODE exhibits a strictly monotonic decrease in both MSE and MAE as $L$ grows. This confirms that the continuous-time integration within the Neural ODE branch acts as an effective noise filter, successfully extracting meaningful low-frequency dynamic patterns from extended historical sequences without suffering from attention dilution.

\section{Limitations and Future Work}
\label{app:limitations}

While CTIA-ODE establishes a robust paradigm for multivariate time series forecasting, several constraints remain that provide avenues for future research. A primary limitation is the assumption of regularly sampled historical observations. The current architecture relies on standard 1D convolutional embeddings, which require the historical sequence $X$ to be sampled at fixed intervals (e.g., hourly or 15-minute). Although the underlying Neural ODE mathematically supports continuous-time evaluation, adapting our encoder to process irregularly sampled data with missing values—such as those found in medical health records—remains a challenge. This would necessitate replacing the current fixed-step encoder with continuous-time variants, such as ODE-RNNs or neural flow models, to capture non-uniform temporal patterns.

Another open challenge is zero-shot generalization. CTIA-ODE is currently trained from scratch on individual datasets, lacking the broad-scale transferability of emerging foundation models (e.g., Chronos or MOIRAI). Investigating how our continuous-time dynamic modules can be pre-trained on diverse, large-scale time series corpora and fine-tuned for unseen downstream tasks represents a promising direction for scale-driven generalization. 

Additionally, our reliance on the 4th-order Runge-Kutta (RK4) solver, while computationally efficient, may not capture the full complexity of highly chaotic or turbulent systems (e.g., high-frequency financial ticks), which often require adaptive-step solvers such as \texttt{dopri5}. However, introducing adaptive solver conditions can complicate training speed and the stability of our loss functions. Future work will explore dynamic thresholding strategies for the Outlier-Robust Hybrid Loss (ORHL) to guarantee mathematical stability under adaptive solvers without sacrificing the computational efficiency demonstrated in this work.

\end{document}